\documentclass{article} % For LaTeX2e
\usepackage{verdi_preprint,times}

\usepackage{amsmath,amsfonts,bm}

\def\eqref#1{equation~\ref{#1}}
\def\1{\bm{1}}

\DeclareMathAlphabet{\mathsfit}{\encodingdefault}{\sfdefault}{m}{sl}
\SetMathAlphabet{\mathsfit}{bold}{\encodingdefault}{\sfdefault}{bx}{n}

\newcommand{\E}{\mathbb{E}}

\usepackage[utf8]{inputenc} % allow utf-8 input
\usepackage[T1]{fontenc}    % use 8-bit T1 fonts
\usepackage{url}            % simple URL typesetting
\usepackage{booktabs}       % professional-quality tables
\usepackage{amsfonts}       % blackboard math symbols
\usepackage{nicefrac}       % compact symbols for 1/2, etc.
\usepackage{microtype}      % microtypography
\usepackage{xcolor}         % colors
\usepackage{graphicx}
\usepackage{tabularx}
\usepackage{subfigure}
\usepackage{mathtools}
\usepackage{amsthm}
\usepackage{amsmath,amssymb,bm}
\usepackage{extarrows}
\usepackage{diagbox}
\usepackage{multirow}
\usepackage{makecell}
\usepackage{algorithm}
\usepackage{algpseudocode}
\usepackage{pgfplots}
\pgfplotsset{compat=1.18}
\usepgfplotslibrary{groupplots}
\usetikzlibrary{arrows.meta} % arrow tips used by Figure~\ref{fig:intro}
\usepackage[hidelinks]{hyperref}       % after float/subfigure to avoid duplicate anchors

\newcommand{\methodname}{VERDI}
\newcommand{\I}{\mathbb{I}}

\newcommand{\cB}{\mathcal{B}}
\newcommand{\cC}{\mathcal{C}}
\newcommand{\LCB}{\operatorname{LCB}}

\providecommand{\E}{\mathbb{E}}

\title{VERDI: Verified Contradictions\\
for Continual World-Model Optimization}

\author{%
Junyu Wu\textsuperscript{*}, Shiqin Nie\textsuperscript{*}, Youyi Kou, Baohua Yin,\\
Qingyu Chen, Jingheng Ma, Shiji Zhou, Hongyong Song, Mingchen Zhuge,\\
Guocai Yao\textsuperscript{\ensuremath{\dagger}},
Tielin Shi\textsuperscript{\ensuremath{\dagger}},
Sen Cui\textsuperscript{\ensuremath{\ddagger,\dagger}},
Changshui Zhang\textsuperscript{\ensuremath{\dagger}}\\[7pt]
{\normalfont\small \textsuperscript{*}Equal contribution\quad
\textsuperscript{\ensuremath{\ddagger}}Project leader\quad
\textsuperscript{\ensuremath{\dagger}}Corresponding author}\\[3pt]
{\normalfont\small\texttt{yaoguocai@pku.edu.cn},\quad\texttt{tlshi@hust.edu.cn}}\\[2pt]
{\normalfont\small\texttt{cuis@mail.tsinghua.edu.cn},\quad\texttt{zcs@mail.tsinghua.edu.cn}}%
}

\hypersetup{
pdfauthor={Junyu Wu, Shiqin Nie, Youyi Kou, Baohua Yin, Qingyu Chen, Jingheng Ma, Shiji Zhou, Hongyong Song, Mingchen Zhuge, Guocai Yao, Tielin Shi, Sen Cui, Changshui Zhang},
pdftitle={VERDI: Verified Contradictions for Continual World-Model Optimization}
}
\begin{document}

\maketitle
\begin{abstract}
World models serve different domains and are tuned to different metrics, from visual fidelity to physical consistency and efficiency, so automating their optimization is valuable.
Their optimization operators---reusable training, representation, sampling, or inference changes---look portable, yet models agreeing on architecture class, size, and baseline score can respond to the same operator in \emph{opposite} directions, so what repairs one degrades its nearest neighbor.
Existing methods supply the search but not the answer: they rank candidates by observational descriptors, assume these predict the same intervention response, and discard failed transfers as noise.
We argue that an operator's effect sign is fixed by the model's response along the direction the operator modifies, not by any descriptor, so similarity is not a sufficient statistic and no fixed description can repair itself.
The repair signal is a \emph{verified contradiction}---nearby models whose opposing outcomes for the same operator are established by probing and by paired experiments under a verifier the search loop cannot edit.
\methodname{} turns this into recursive self-optimization through shared-probe Optimization Fingerprints, IRG-guided retrieval, frozen-verifier settlement, and probe admission gated by held-out replay.
Across nine backbones, 27 leave-one-backbone-out campaigns, and 216 transfer decisions, \methodname{} predicts effect signs with 83\% accuracy, reaches the first verified positive in 3.6 trials and 96 GPU-h (68\% and 69\% below cold start), and lowers negative-transfer risk from 0.34 to 0.06 at 47\% coverage.
Ablations confirm that the geometry, not permissive acceptance, does the work, and a small-scale Franka Panda study gains 17.5 percentage points on two tasks.
\textbf{Project page:} \url{https://verdiwm-anno.github.io/}
\end{abstract}

% \begin{center}
%   
%   \small\textbf{Project page:} \url{https://verdiwm-anno.github.io/}
% \end{center}
% 

% The template default is 6pt; a smaller body paragraph gap avoids oversized
% whitespace while preserving visible paragraph boundaries.
% \setlength{\parskip}{1pt plus 0.2pt minus 0.2pt}

\section{Introduction}
\label{sec:intro}

World models predict how environments evolve and are used for planning, simulation, and control~\citep{ha2018worldmodels,hafner2020dream,hafner2025mastering}.
They serve different domains---robotics, driving, interactive simulation---so different backbones are tuned to different metrics, from visual fidelity to physical consistency or efficiency~\citep{nvidia2025cosmos,guo2025ctrlworld,googdeepmind2025genie3,han2026wam}, and each still needs its own improvement campaign.
Automated, continual world-model optimization is therefore valuable: it turns a per-model effort into experience that compounds.
We call a reusable optimization operation an \emph{operator}: an executable change to training, representation, sampling, or inference.
Operators look portable, yet models that agree on architecture class, parameter count, and baseline scores can respond to the same operator in \emph{opposite} directions, so what repairs one degrades its nearest neighbor.
We therefore ask: \emph{how can optimization experience accumulate when the models that look most alike can call for opposite interventions?}

\begin{figure}[t]
  \setlength{\abovecaptionskip}{0pt}
  \setlength{\belowcaptionskip}{0pt}
    \centering
    % Insight-centric Figure 1 (TikZ). To restore the previous Canva panel,
    % comment the \input line and uncomment the \includegraphics line below.
    %\input{figure/introduct1.pdf}
     \includegraphics[width=\textwidth]{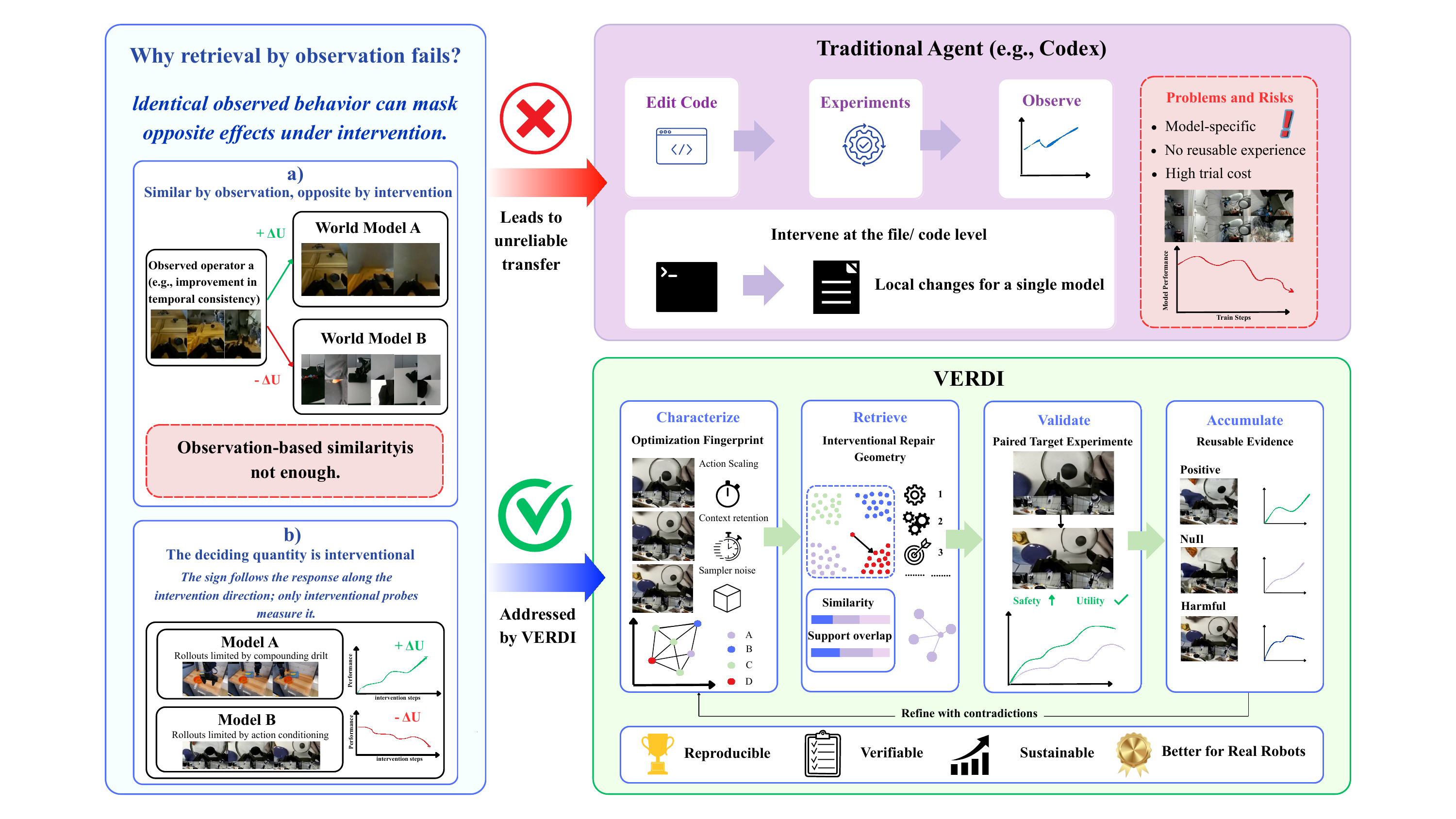}
    \caption{\textbf{From observational similarity to reusable repair evidence.} Left: (a) observationally similar models can respond oppositely to the same operator; (b) its effect depends on the interventional response along the direction it modifies. Upper right: a conventional agent edits code and evaluates changes within one model. Lower right: VERDI characterizes models with shared probes, retrieves operators by response geometry, validates paired target experiments under a frozen verifier, and accumulates positive, null, and harmful outcomes for reuse. Verified contradictions guide probe refinement, with new probes admitted only when held-out replay improves selection and calibration.}
    \label{fig:intro}
  
\end{figure}

Existing methods supply the search but not the answer.
Backbone-specific methods improve one model at a time~\citep{guo2025ctrlworld,nvidia2025cosmos,zhu2025irasim}; research agents automate proposing, implementing, and evaluating such improvements~\citep{lu2024aiscientist,huang2024mlagentbench,novikov2025alphaevolve,zhang2025dgm}; warm-started transfer reuses priors across related tasks~\citep{feurer2015initializing,bai2023transferbo,achille2019task2vec,chen2022optformer}.
All of them rank candidates by observational descriptors---architecture, metadata, embeddings, scores---and assume these predict the same intervention response, so a failed transfer is discarded as noise rather than as evidence about the description that failed.

We argue that the bottleneck is that these descriptors are observational: whether an operator helps is decided by the model's \emph{response along the direction that the operator modifies}, relative to the error that dominates its rollouts.
Two models can match on every descriptor yet differ in what limits them, so the same operator repairs one and degrades the other.
Similarity is thus not a sufficient statistic for intervention effects, and no fixed description can repair itself, because it carries no signal about the distinction it is missing.
That signal is observable in a \emph{verified contradiction}: two nearby models whose opposing outcomes for the same operator survive cheap inference-time probing and paired experiments under a verifier that the search loop cannot edit.

Continual optimization should therefore co-evolve two objects in one loop: the operator library, and the behavioral description that decides where each operator applies.
Models should be characterized by interventional responses, not metadata; candidates should be settled by experiment, not prediction, so that null and harmful outcomes become evidence; and a verified contradiction should trigger the probe that best explains it, admitted only if held-out replay improves selection.
Each batch of campaigns then consumes settled outcomes and returns a better geometry, instead of starting cold.

We instantiate this view in \methodname{}, a four-stage loop (Figure~\ref{fig:intro}).
\emph{Characterize} forms an \emph{Optimization Fingerprint} from shared inference-time probes.
\emph{Retrieve} ranks executable operators with the Interventional Repair Geometry (IRG) over an Operator registry and Operator outcome archive.
\emph{Validate} runs paired target experiments that a frozen verifier---not the proposal or the search loop---settles.
\emph{Accumulate} records each outcome with its support, uncertainty, and lineage, and admits a new probe only when held-out replay improves selection and calibration.

Experiments on ACWM-Phys~\citep{xue2026acwmphys} and nine pretrained backbones test both claims.
Selected operators raise Ctrl-World motion smoothness from 68.75 to 81.53 and COSMOS3 simulated control by 12--24 points, and a small-scale Franka Panda study gains 17.5 percentage points.
Across 27 leave-one-backbone-out campaigns and 216 transfer decisions, \methodname{} predicts effect signs with 83\% accuracy, reaches the first verified positive in 3.6 trials and 96 GPU-h, 68\% and 69\% below cold start, and holds negative-transfer risk at 0.06 over 47\% coverage against 0.34 under unconditional reuse.
Ablations show that the geometry does the work: raw responses or learned embeddings reach only 0.71 and 0.74 sign accuracy, and removing IRG gives 4.8 trials and 0.18 risk.
Under a matched 200 GPU-h allowance, \methodname{} completes 24/27 campaigns in 3.6 trials, versus 20/27 for Codex (GPT-6).
Our contributions are threefold:
\begin{itemize}
    \item \textbf{A paradigm for continual world-model optimization.} Verified opposing outcomes between models that a similarity description places close together reveal a missing behavioral distinction, so operators and the geometry that predicts where they apply should co-evolve.

    \item \textbf{A response-aware framework.} \methodname{} realizes this as a four-stage loop of Optimization Fingerprints, IRG-guided retrieval, frozen-verifier validation, and an outcome archive whose probe refinement must improve held-out selection.

    \item \textbf{A benchmark and verification protocol.} Leave-one-backbone-out evaluation over nine backbones, 27 campaigns, and 216 transfer decisions, with archive replay and a planted-fault audit, verifies both claims.
\end{itemize}

\section{Related Work}
\label{sec:related-work}
\paragraph{World-model evaluation.}
Evaluation now extends beyond visual fidelity to dynamics, action fidelity, and embodied utility.
EWMBench assesses scene, motion, and semantics~\citep{yue2025ewmbench}; WorldArena and WorldArena2.0 combine video metrics with embodied tasks~\citep{shang2026worldarena,shang2026worldarena2}.
GigaWorld analyzes relations among evaluation metrics~\citep{gigaai2025gigaworld}.
These benchmarks define outcomes; VERDI asks which operator changes them for a given model and whether its effect transfers across backbones.

\paragraph{World-model improvement.}
World models have advanced through latent imagination for control~\citep{hafner2025mastering}, controllable manipulation dynamics~\citep{guo2025ctrlworld,zhu2025irasim}, and large video or world-action backbones~\citep{nvidia2025cosmos,han2026wam}.
These works improve individual models through architecture, training, or conditioning choices.
VERDI studies how such interventions become reusable evidence across backbones by recording their measured effects, support conditions, and validity boundaries.

\paragraph{Research and self-improving agents.}
Research agents automate code changes and experimental search~\citep{lu2024aiscientist,novikov2025alphaevolve}, including robot-policy improvement loops~\citep{xie2026enpire,li2026roboclaw}.
Warm-started optimization, task embeddings, and optimization-history models guide new decisions from prior tasks~\citep{feurer2015initializing,achille2019task2vec,chen2022optformer}.
These lines of work automate search or transfer optimization priors, but do not directly model how an operator affects heterogeneous world-model backbones.
VERDI uses validated source--target effects to select operators and refines its response geometry when nearby models with comparable support yield opposing effects.

\section{Method}
\label{sec:method}

\subsection{Problem Setting}
\label{sec:problem-setting}

We study repeated campaigns for improving pretrained world models. In campaign $x$, a model $M_x$ produces future observations, and, when supported by its interface, actions or task events from a context and optional control inputs. The campaign $\mathcal{C}_x=(M_x,\mathcal{D}_x,\mathcal{G}_x,V_x,B_x)$ fixes the permitted data $\mathcal{D}_x$, optimization goal $\mathcal{G}_x$, verifier $V_x$, and nominal experimental budget $B_x$. The goal specifies a utility $U_{\mathcal{G}_x}$, protected outcomes that must remain within declared tolerances, and a minimum meaningful gain $\tau_{\mathcal{G}_x}$. This formulation covers both action-conditioned world models and world-action models; backbone-specific details are handled by adapters.

The object we seek to reuse is an operator $a$: an executable change to training, representation, sampling, or inference with a specified model hook, dose, schedule, and preconditions. Its value is an empirical target-side outcome, not the fact that it worked once. Let $M_x^{(a)}$ be the model after applying $a$, and $M_x^{(0)}$ its baseline control. The campaign utility and its paired gain are
\begin{equation}
U_x(a)=U_{\mathcal{G}_x}\!\left(M_x^{(a)};\mathcal{D}_x,V_x\right),
    \qquad \Delta U_x(a)=U_x(a)-U_x(0).
    \label{eq:differential-repair}
\end{equation}
Both arms use matched initialization, data, evaluation contexts, rollout seeds, and the same frozen verifier. Protected outcomes are checked separately against their declared tolerances. An operator outcome is positive only when the lower confidence bound on $\Delta U_x(a)$ exceeds $\tau_{\mathcal{G}_x}$ and every protected check passes. A utility gain cannot compensate for a protected regression. Throughout, a \emph{trial} is one candidate operator attempt; compilation, screening, confirmation, and additional rollout seeds are events within that attempt.

\subsection{Method Overview}
\label{sec:method-overview}

\begin{figure}[t]
  \setlength{\abovecaptionskip}{0pt}
  \setlength{\belowcaptionskip}{0pt}
    \centering
    \includegraphics[width=\textwidth]{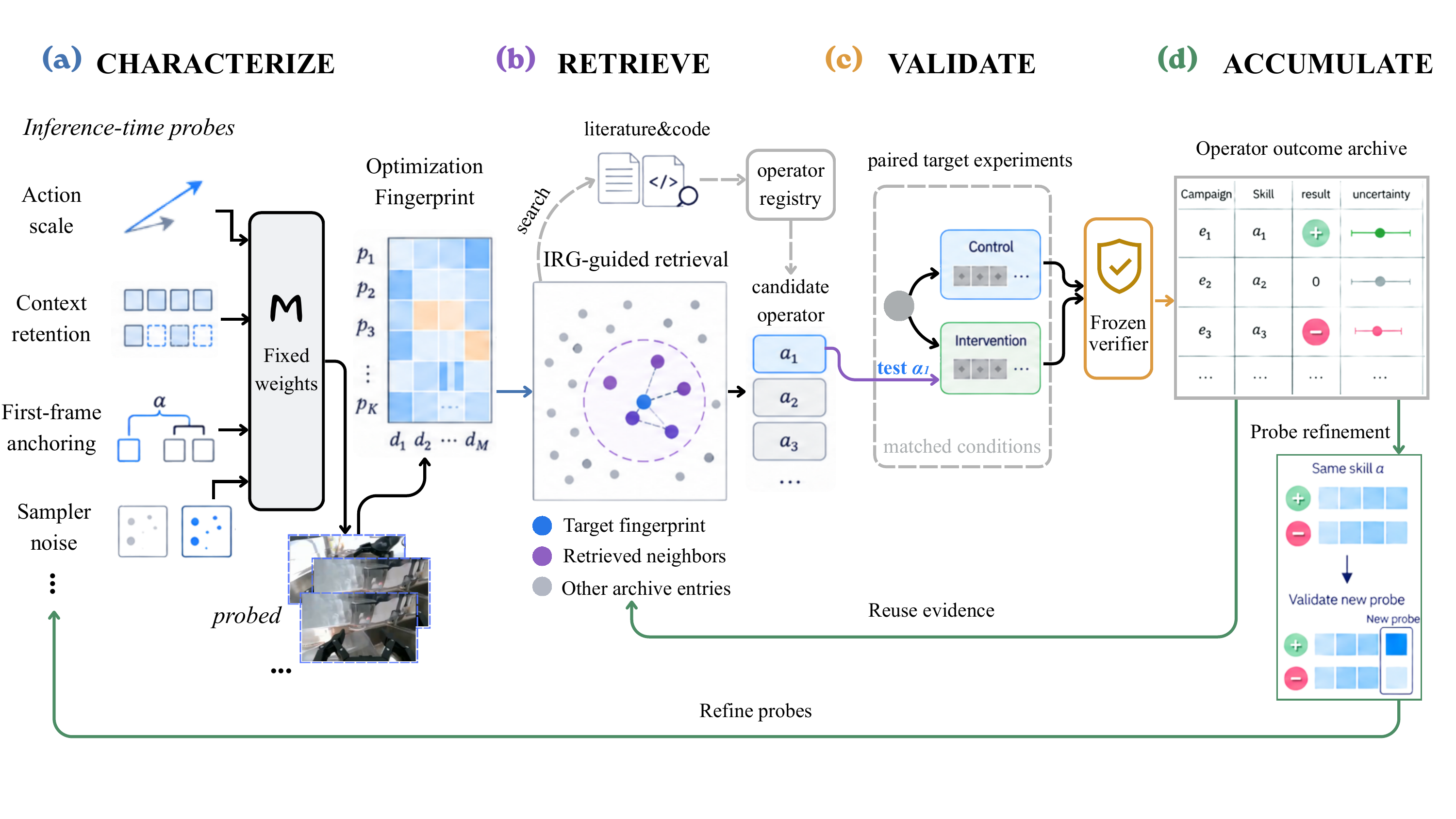}
    \caption{\textbf{VERDI's continual optimization loop.}
    \emph{Characterize} measures shared inference-time responses to build an
    Optimization Fingerprint. \emph{Retrieve} uses IRG-guided retrieval to
    compare the fingerprint with archived models and may search literature or code for a new operator.
    \emph{Validate} runs paired target experiments under a frozen verifier, while
    \emph{Accumulate} writes settled positive, null, harmful, and interaction
    outcomes to an Operator outcome archive. The feedback path admits a new probe only
    after held-out checks.}
    \label{fig:Verdi_overview}
  
\end{figure}

VERDI follows four stages (Figure~\ref{fig:Verdi_overview}).
\emph{Characterize} measures responses to a shared set of inference-time probes and forms an Optimization Fingerprint.
\emph{Retrieve} uses IRG-guided retrieval to rank operators using the Operator registry and Operator outcome archive and to guide external search when no suitable record exists.
\emph{Validate} tests candidates through paired screening and held-out confirmation on the target model.
\emph{Accumulate} records the settled operator outcome and uses reliable opposing outcomes among nearby models to decide whether an additional probe is useful.

For illustration, suppose models A and B respond similarly to existing
probes, but a temporal operator helps A and harms B under paired validation.
This disagreement suggests a missing distinction. VERDI proposes a
history-sensitive probe, remeasures both models, and retains the probe only
if held-out replay improves selection.

A fixed LLM executor searches external sources and implements operators through
backbone adapters with implementation receipts. It cannot change the verifier
or decide which outcomes enter the archive. In transfer evaluation, the Operator
outcome archive and Operator registry are frozen before target evaluation; newly
discovered operators enter later versions.

\subsection{Characterize: Building the Optimization Fingerprint}
\label{sec:fingerprint}

We probe a model at inference time, before choosing a training or architecture change. Each shared probe $q\in\mathcal{Q}$ makes a small, reversible change of dose $\alpha$ to a behavioral input or inference setting: for example, action magnitude, retained temporal context, first-frame anchoring, or sampler noise. Adapters translate the same probe contract to each backbone without updating weights or changing the optimization goal. For diagnostic and validity measurements $z_k$ on matched contexts $\xi$ and seeds $\omega$, we estimate
\begin{equation}
    J_x[k,q]=\mathbb{E}_{\xi,\omega}\!\left[
    \frac{z_k(M_x^{q,\alpha};\xi,\omega)-z_k(M_x;\xi,\omega)}{\alpha}
    \right].
    \label{eq:response-jacobian}
\end{equation}
Each column of $J_x$ records one probe's response. Pairing reduces sampling
noise, and bootstrap estimates give uncertainty $\Sigma_x$. Action scaling
probes action sensitivity; context retention probes history dependence.
These diagnostics guide candidate operator selection; actual outcomes still require
paired target-side experiments.

We call $\phi_x=(J_x,G_x,\Sigma_x,c_x)$ the \emph{Optimization Fingerprint}. Here $G_x=J_x^\top W_xJ_x+\lambda I$ weights standardized diagnostic outcomes according to the goal and regularizes weak probe directions; $c_x$ records which probe and operator hooks the adapter supports. The fingerprint provides a common behavioral description across backbones while keeping measurement uncertainty and implementation compatibility explicit. Probes measure the model's responses; operators are the optimization operations subsequently selected and tested.

\subsection{Retrieve: IRG-Guided Operator Selection}
\label{sec:retrieve}
\label{sec:method-irg}

The Interventional Repair Geometry (IRG) organizes models and their historical
operator outcomes using responses to shared inference-time probes. It compares
compatible Optimization Fingerprints, with goal-specific weighting and a penalty
for mismatched evaluation support. Two models are close when the same controlled
perturbations change relevant outcomes in similar ways, even if their
architectures differ. Fingerprint uncertainty enters confidence estimates and
abstention.

The Operator registry stores executable operator definitions and their implementation
requirements; the Operator outcome archive records what happened when those operators
were applied. IRG-guided retrieval ranks candidate operators using outcomes from
compatible neighbors. The target fingerprint, failure coordinates, supported
hooks, and goal also guide external search for additional candidates. A
discovered candidate is compiled and evaluated by the same target-side screen
and verifier; retrieval alone never assigns a target outcome. The archive and
registries are frozen before held-out outcomes are observed, and newly discovered
operators belong to a later version.

The Operator outcome archive links each operator and source fingerprint to its
settled gain, protected outcomes, support, uncertainty, cost, verifier result,
and artifact lineage. It retains positive, null, harmful, and jointly tested
interaction outcomes. For target $t$ and operator $a$, only records compatible with
its goal, support, and adapter capabilities contribute to the gain estimate:
\begin{equation}
    \widehat{\Delta U}_t(a)=
    \frac{\sum_{s\in\mathcal{N}_t(a)}w_{ts}(a)\,r_s^U(a)}
         {\sum_{s\in\mathcal{N}_t(a)}w_{ts}(a)},\qquad
    w_{ts}(a)\propto
    \frac{\rho_s(a)\kappa_{ts}}
    {\epsilon+\operatorname{dist}_{\mathrm{IRG}}(\phi_t,\phi_s)}.
    \label{eq:effect-retrieval}
\end{equation}
Here $r_s^U(a)=\Delta U_s(a)$ is the source gain, $\rho_s(a)$ measures
evidence quality, and $\kappa_{ts}$ measures support overlap. The selector
allocates trials by an upper-confidence gain-per-cost score; separate
compatibility checks determine transfer priority. These estimates rank
hypotheses, while target trials establish effects.

\subsection{Validate: Paired Target Experiments}
\label{sec:validate}
\label{sec:transfer}

Before a retrieved operator receives transfer priority, VERDI checks that it compiles for the target, source and target supports overlap, enough credible source evidence exists, probe responses align, nearby effect signs agree, and a calibrated lower bound on predicted gain clears the goal threshold. This check prevents weak or conflicting evidence from being treated as a transferable result. A candidate that fails it can still be explored from a cold start if it compiles locally. Protected outcomes are measured on the operator-applied target model.

Each selected operator first receives a paired, low-cost screen against the matched control. Promising candidates then undergo held-out confirmation with longer rollouts, more optimization steps, and additional seeds. The frozen verifier settles the result using the campaign's predeclared utility threshold and protected tolerances. A positive record requires a lower confidence bound on target gain above $\tau_{\mathcal{G}_t}$ and all protected checks to pass. Valid confirmed null and harmful operator outcomes are also recorded with their uncertainty; compilation failures, invalid runs, and unconfirmed screens do not become settled outcome evidence. These paired target experiments establish whether the operator improves the target model.

\subsection{Accumulate: Recording Outcomes and Refining Probes}
\label{sec:evolution}

After validation, VERDI writes the settled operator outcome together with its support, uncertainty, verifier decision, cost, operator version, and implementation receipt to the Operator outcome archive. Positive records provide future candidates, while null and harmful records mark limits of applicability. Jointly measured combinations receive interaction records rather than being attributed to either operator alone. Source--target tests update the evidence available for future retrieval without rewriting the original source records.

The archive also reveals when its behavioral description is insufficient. In the example above, models A and B are close in IRG but show reliable opposing outcomes for the same temporal operator. The current fingerprint may therefore omit a distinction relevant to operator selection; an isolated failure without a comparable nearby source does not establish this. VERDI uses the disagreement context and available adapter hooks to propose a history-sensitive probe, and retains it only if held-out replay improves operator selection and calibration checks pass. Probe refinement updates IRG for later campaigns, while every claimed operator outcome remains grounded in a verifier-controlled target experiment.

\section{Experiments}
\label{sec:experiments}

We evaluate model-specific improvements, compare transfer and operator-search
efficiency, and then isolate the contributions of selection, probe evolution,
and verification through ablations.

\subsection{Experimental Protocol}
\label{sec:exp-protocol}

Campaigns freeze checkpoints, splits, contexts, seeds, metrics, validity gates,
registry snapshots, and nominal budgets. Baseline and operator-applied arms share
initialization, data, schedules, and evaluators. The LOBO study covers nine
backbones with three seeds each (27 campaigns, 216 transfer decisions;
App.~\ref{tab:shared-prior-archive}), excluding each target's records
from its source archive. Transfer and controlled selector/scaffold comparisons
use this study, as do archive-prefix replay and external-system comparisons.
Confidence intervals resample the nine backbone clusters with their seeds
retained; separate case studies state their own evaluation scope.
We distinguish cost to the
first verified positive, $C_{\mathrm{first+}}$, from full search cost,
$C_{\mathrm{search}}$ (App.~\ref{app:cost-ledger}). A 512-step screen orders
candidates; the official 50-step evaluation and held-out confirmation settle
effects. Protected metrics are checked separately from utility
(App.~\ref{app:extended-protocol}).

\subsection{Targeted Outcomes of Model-Specific Optimization}
\label{sec:exp-acwm}
\label{sec:exp-backbone-slices}
\label{sec:exp-real-robot}

We evaluate visual quality, physical-interaction prediction, and real-robot
control to distinguish improvements in generated observations from downstream
action outcomes.

\begin{figure}[!htb]
  \setlength{\abovecaptionskip}{0pt}
  \setlength{\belowcaptionskip}{0pt}
  \centering
  \begin{minipage}[t]{0.49\linewidth}
    \centering\vspace{0pt}
    \textbf{(a) Ctrl-World: latent-motion prior}\par
    \includegraphics[width=\linewidth,trim=174 69 178 51,clip]{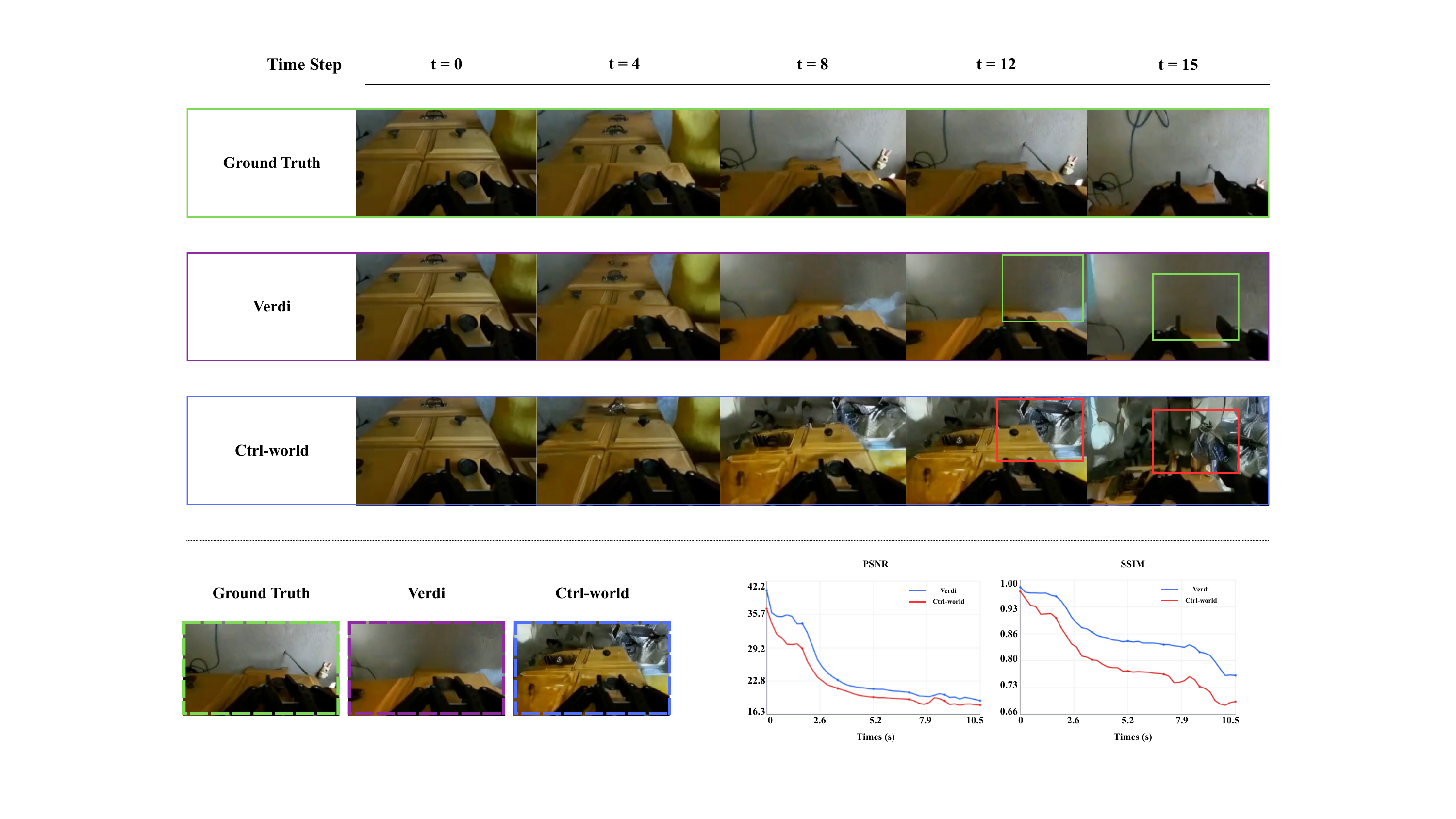}
  \end{minipage}\hfill
  \begin{minipage}[t]{0.49\linewidth}
    \centering\vspace{0pt}
    \textbf{(b) COSMOS3: RL + self-forcing}\par
    \includegraphics[width=\linewidth,trim=142 11 142 14,clip]{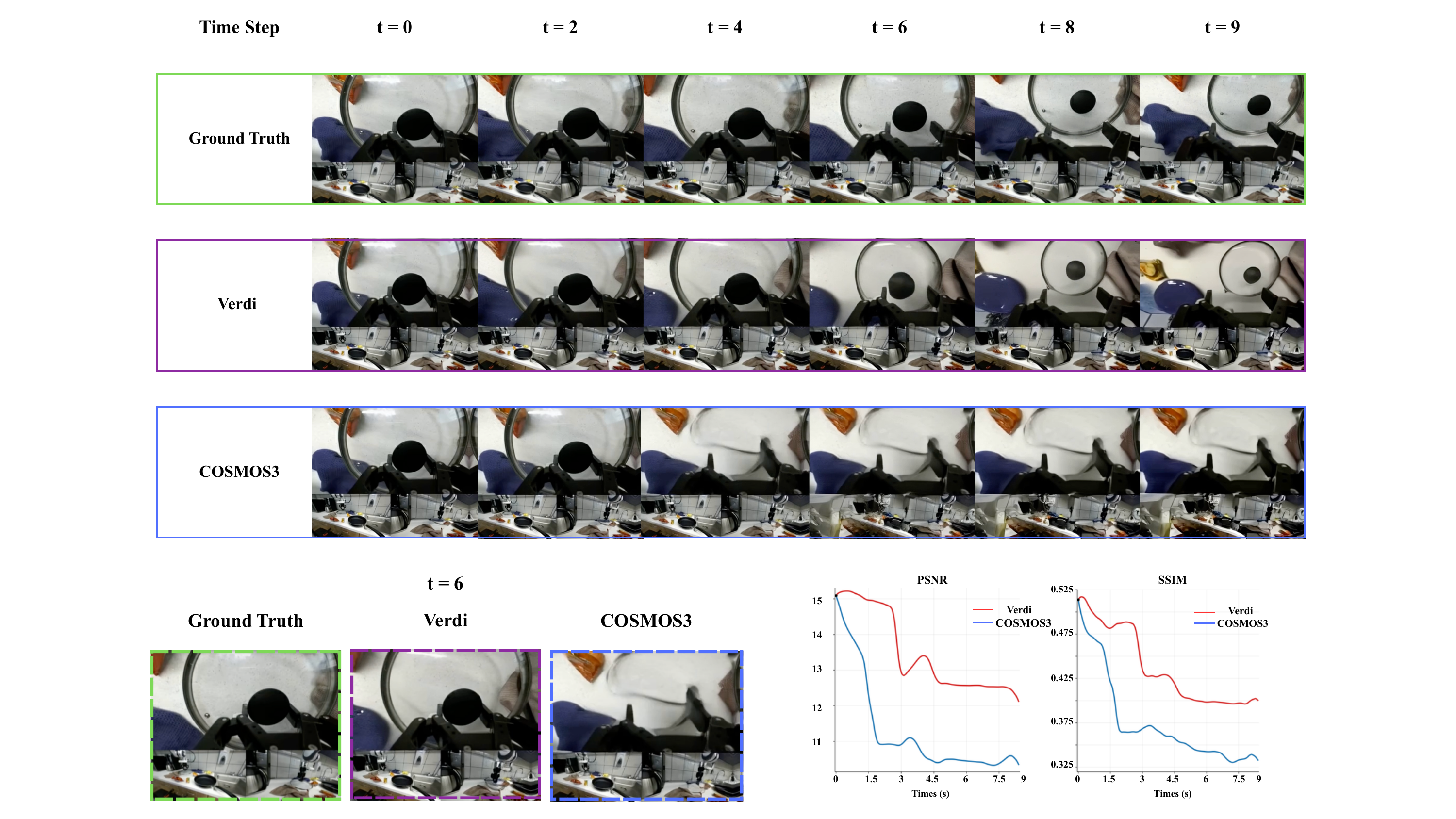}
  \end{minipage}
  \caption{\textbf{Visual quality across backbones.} Ground truth, baseline,
  and operator-applied rollouts, with paired PSNR/SSIM curves. (a) Ctrl-World's
  latent-motion operator. (b) COSMOS3 with RL + self-forcing on ``loosen the pot
  handle'' (DROID~\citep{khazatsky2024droid}).}
  \label{fig:ctrl-world-main}
  \label{fig:cosmos3-long-horizon}
  
\end{figure}

\paragraph{Visual quality.}
On Ctrl-World, VERDI selects a latent-motion prior that reduces rollout drift
(Figure~\ref{fig:ctrl-world-main}a). Four primary coordinates improve, including
motion smoothness from 68.75 to 81.53; subject consistency falls by 0.96 within
tolerance (Table~\ref{tab:ctrl-world-main}). A protected dynamic-degree
regression on open-drawer limits these gains to descriptive coordinate
improvements, not a verified gain over the full scope (App.~\ref{app:ctrl-world}).
On COSMOS3, a combined reinforcement-learning (RL) and self-forcing operator
reduces subject and background drift in long rollouts, while the complementary
WAM and real-robot results below test whether the operator reaches downstream
control (App.~\ref{app:cosmos-family}).

\begin{table}[!htb]
  \setlength{\abovecaptionskip}{0pt}
  \setlength{\belowcaptionskip}{0pt}
  \centering
  \caption{\textbf{Model-specific optimization results.} Backbones match
  Figure~\ref{fig:ctrl-world-main}: (a) Ctrl-World visual metrics;
  (b) COSMOS3 simulation success and action diagnostics under the combined
  RL + self-forcing operator. Success decay is the 8-step to 14-step task gap in
  percentage points.}
  \label{tab:ctrl-world-main}
  \small
  \begin{minipage}[t]{0.49\linewidth}
    \centering\vspace{0pt}
    \textbf{(a) Ctrl-World: visual quality}\par
    \setlength{\tabcolsep}{3pt}
    \begin{tabular}{@{}lrr@{}}
      \toprule
      Metric $\uparrow$ & Base & VERDI \\
      \midrule
      Semantic alignment & 90.70 & \textbf{91.30} \\
      Depth accuracy & 93.28 & \textbf{96.16} \\
      Flow score & 26.54 & \textbf{30.35} \\
      Motion smoothness & 68.75 & \textbf{81.53} \\
      Subject consistency & \textbf{84.31} & 83.35 \\
      \bottomrule
    \end{tabular}
  \end{minipage}\hfill
  \begin{minipage}[t]{0.49\linewidth}
    \centering\vspace{0pt}
    \textbf{(b) COSMOS3: simulated control}\par
    \setlength{\tabcolsep}{1.5pt}
    \begin{tabular}{@{}lrr@{}}
      \toprule
      Diagnostic & COSMOS3 & VERDI \\
      \midrule
      Pick-and-place success $\uparrow$ & 62\% & \textbf{74\%} \\
      Drawer opening success $\uparrow$ & 41\% & \textbf{59\%} \\
      Peg insertion success $\uparrow$ & 28\% & \textbf{52\%} \\
      Expert-action L2 error $\downarrow$ & 0.084 & \textbf{0.061} \\
      Success decay $\downarrow$ & 34\,pts & \textbf{22\,pts} \\
      \bottomrule
    \end{tabular}
  \end{minipage}
  
\end{table}

\paragraph{Physical consistency.} Table~\ref{tab:acwm-phys-consistency} reports operator-applied minus baseline changes
on the eight ACWM-Phys environments under the frozen evaluator. PSNR and
SSIM increase, while masked MSE decreases, in every environment. The largest
PSNR gain occurs on \texttt{robot\_arm} (+1.08 dB); \texttt{cloth\_move}
shows the largest SSIM gain (+0.0099) and masked-MSE reduction
($-9.040\times10^{-3}$).

\begin{table}[!htb]
  
  \setlength{\abovecaptionskip}{0pt}
  \setlength{\belowcaptionskip}{0pt}
  \centering
  \caption{\textbf{ACWM-Phys physical-consistency summary.}
  Entries are operator-applied minus baseline changes under the frozen evaluator.
  Higher PSNR and SSIM and lower masked MSE (mMSE) are better.
  $\Delta$mMSE is reported in units of $10^{-3}$.}
  \label{tab:acwm-phys-consistency}
  \begin{tabular}{@{}lrrr@{}}
    \toprule
    Environment & $\Delta$PSNR (dB) $\uparrow$ & $\Delta$SSIM $\uparrow$ &
    $\Delta$mMSE ($10^{-3}$) $\downarrow$ \\
    \midrule
    push\_cube  & +0.15 & +0.0002 & -0.284 \\
    stack\_cube & +0.03 & +0.0020 & -0.531 \\
    push\_rope  & +0.02 & +0.0008 & -0.151 \\
    cloth\_move & +0.94 & +0.0099 & -9.040 \\
    push\_sand  & +0.64 & +0.0010 & -0.783 \\
    pour\_water & +0.85 & +0.0019 & -1.718 \\
    robot\_arm  & +1.08 & +0.0098 & -2.230 \\
    reacher     & +0.06 & +0.0002 & -0.119 \\
    \bottomrule
  \end{tabular}
  
\end{table}

\begin{figure}[!htb]
  \setlength{\abovecaptionskip}{0pt}
  \setlength{\belowcaptionskip}{0pt}
  \centering
  \includegraphics[width=0.70\textwidth,trim=0 65 0 15,clip]{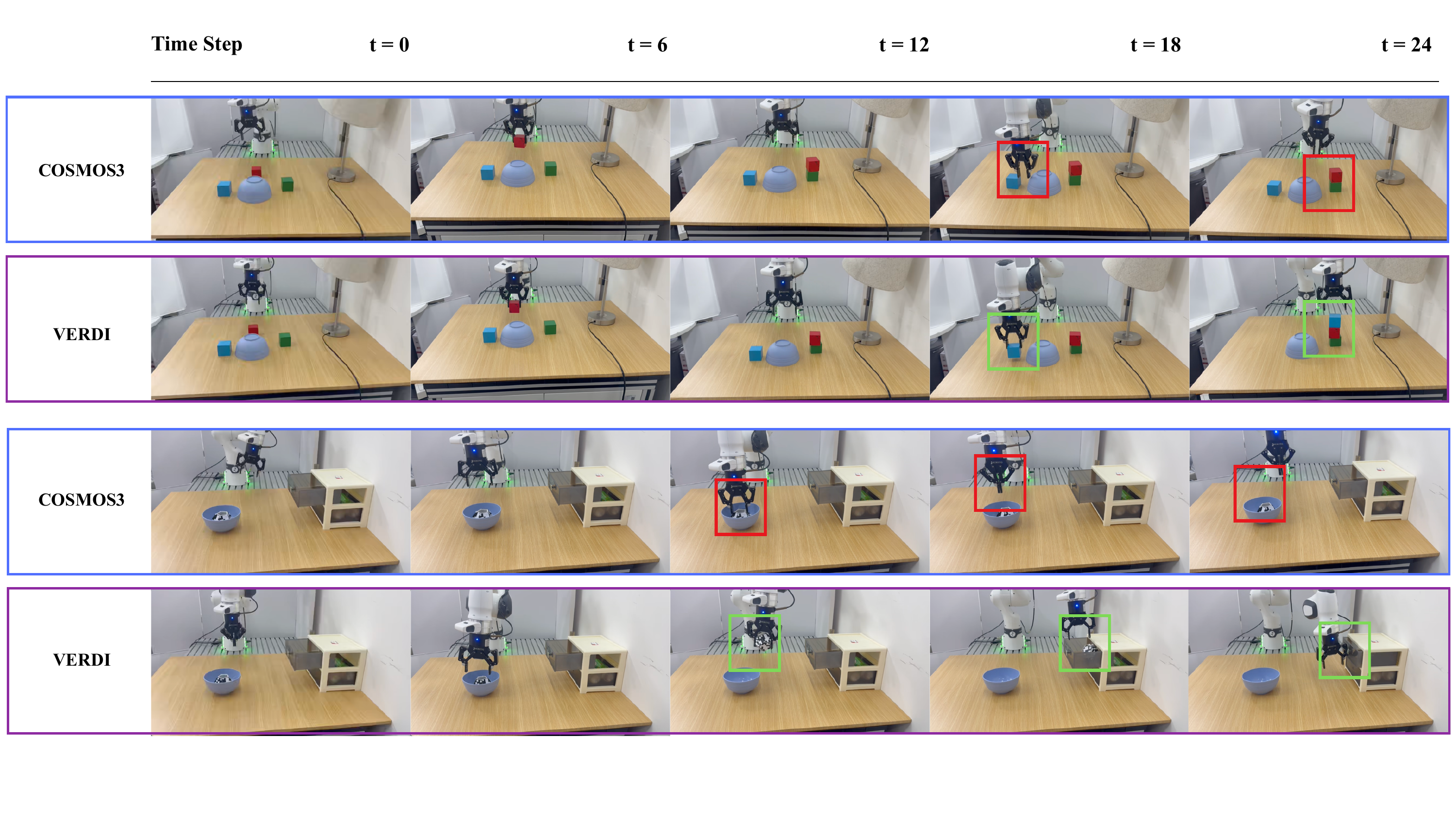}
  \caption{\textbf{Real-robot control.} Baseline COSMOS3 (blue) and COSMOS3
  with RL + self-forcing (purple) on pick-and-place and drawer opening, with
  matched initial states and instructions. Boxes mark failed (red) and
  successful (green) actions. Additional qualitative rollouts on the AgileX Cobot Magic dual-arm robot are provided in Appendix~\ref{app:backbone-experiments}, Figure~\ref{fig:songling}.}
  \label{fig:real-robot}
  
\end{figure}

\paragraph{Real-world robot control.}
We evaluate the operator-applied COSMOS3 world-action model (WAM), using the combined
RL + self-forcing operator, separately on a
Franka Panda with eight-step action chunks and re-observation
(Figure~\ref{fig:real-robot}).
Success rises from 11/20 to 14/20 for pick-and-place and from 7/20 to 11/20
for drawer opening: +17.5 percentage points on average. Marginal 95\% Wilson
intervals are [34,74]\% to [48,85]\% and [18,57]\% to [34,74]\%, respectively.
These small-sample results suggest downstream benefit.

\subsection{Cross-Backbone Transfer and Search Efficiency}
\label{sec:exp-archive-growth}
\label{sec:exp-transfer}

Useful individual operators do not establish that experience transfers.
Chronological-prefix replay tests this on the same held-out campaigns:
as the archive grows, sign accuracy rises from 0.61 to 0.83, trials fall from
7.8 to 3.6, and negative-transfer risk falls from 0.18 to 0.06 while coverage
rises from 0.22 to 0.47 (Figure~\ref{fig:archive-growth}).

\begin{figure}[!htb]
  \setlength{\abovecaptionskip}{0pt}
  \setlength{\belowcaptionskip}{0pt}
  
  \centering
  
  \begingroup
\begin{tikzpicture}
\begin{groupplot}[
  group style={group size=3 by 1,horizontal sep=1.0cm},
  width=.30\textwidth,height=.20\textwidth,
  xmin=4,xmax=20,xtick={4,12,20},
  grid=major,tick label style={font=\scriptsize},
  label style={font=\scriptsize},title style={font=\scriptsize}]
\nextgroupplot[title={(a) Sign accuracy},ylabel={Accuracy $\uparrow$},ymin=.45,ymax=.95]
\addplot+[blue,thick,mark=*,error bars/.cd,y dir=both,y explicit]
coordinates {(4,.61) +- (0,.08) (8,.67) +- (0,.07) (12,.73) +- (0,.06) (16,.79) +- (0,.05) (20,.83) +- (0,.04)};
\nextgroupplot[title={(b) Trials to first positive},ylabel={Trials $\downarrow$},ymin=2,ymax=13]
\addplot+[blue,thick,mark=*,error bars/.cd,y dir=both,y explicit]
coordinates {(4,7.8) +- (0,.7) (8,6.4) +- (0,.6) (12,5.2) +- (0,.5) (16,4.3) +- (0,.4) (20,3.6) +- (0,.3)};
\addplot+[gray,dashed,thick] coordinates {(4,11.2)(20,11.2)};
\nextgroupplot[title={(c) Negative-transfer risk},ylabel={Risk $\downarrow$},ymin=0,ymax=.42]
\addplot+[blue,thick,mark=*,error bars/.cd,y dir=both,y explicit]
coordinates {(4,.18) +- (0,.06) (8,.14) +- (0,.05) (12,.11) +- (0,.04) (16,.08) +- (0,.03) (20,.06) +- (0,.02)};
\addplot+[gray,dashed,thick,mark=square*] coordinates {(4,.27)(8,.29)(12,.31)(16,.33)(20,.34)};
\end{groupplot}
\end{tikzpicture}
\endgroup
\caption{\textbf{Chronological archive replay.} Frozen prefixes evaluated on
  27 campaigns from nine backbones; bars are descriptive 95\%
  backbone-cluster bootstrap intervals.}
  \label{fig:archive-growth}
  
\end{figure}

With the full archive, VERDI identifies the first verifier-confirmed positive
operator outcome after 3.6 trials and 96 GPU-h on average, compared with 11.2 trials
and 312 GPU-h for cold-start search. Across 216 held-out transfer decisions,
its compatibility gate achieves 47\% coverage with 6\% negative-transfer risk,
versus 34\% risk at full coverage for unconditional reuse. Coverage measures
the gate's acceptance rate, while risk is conditional on accepted transfers.
Per-backbone breakdowns appear in App.~\ref{app:coverage-risk}.

\begin{figure}[!htb]
  \setlength{\abovecaptionskip}{0pt}
  \setlength{\belowcaptionskip}{0pt}
  
  \centering
  \begingroup
\definecolor{transfercold}{HTML}{7A8594}
\definecolor{transferreuse}{HTML}{B97836}
\definecolor{transferverdi}{HTML}{167C80}
\pgfplotsset{transferpanel/.style={
  width=.305\linewidth,height=2.9cm,
  xmin=.55,xmax=3.45,xtick={1,2,3},
  xticklabels={Cold,Uncond.,VERDI},
  tick label style={font=\scriptsize},label style={font=\scriptsize},
  title style={font=\small},axis x line*=bottom,axis y line*=left,
  axis line style={black!40},tick style={black!40},
  ymajorgrids,grid style={black!9},clip=false}}
\begin{tikzpicture}[baseline=(current bounding box.north)]
\begin{axis}[transferpanel,title={(a) Trials to first positive $\downarrow$},
  ymin=0,ymax=14,ytick={0,4,8,12}]
\addplot[only marks,mark=*,mark size=2.4pt,transfercold] coordinates {(1,11.2)};
\addplot[only marks,mark=*,mark size=2.4pt,transferreuse] coordinates {(2,4.1)};
\addplot[only marks,mark=*,mark size=2.7pt,transferverdi] coordinates {(3,3.6)};
\node[font=\tiny,text=transfercold,anchor=south] at (axis cs:1,11.5) {11.2};
\node[font=\tiny,text=transferreuse,anchor=south] at (axis cs:2,4.4) {4.1};
\node[font=\tiny,text=transferverdi,anchor=south] at (axis cs:3,3.9) {3.6};
\end{axis}
\end{tikzpicture}\hfill
\begin{tikzpicture}[baseline=(current bounding box.north)]
\begin{axis}[transferpanel,title={(b) Measured GPU-h $\downarrow$},
  ymin=0,ymax=370,ytick={0,100,200,300}]
\addplot[only marks,mark=*,mark size=2.4pt,transfercold] coordinates {(1,312)};
\addplot[only marks,mark=*,mark size=2.4pt,transferreuse] coordinates {(2,118)};
\addplot[only marks,mark=*,mark size=2.7pt,transferverdi] coordinates {(3,96)};
\node[font=\tiny,text=transfercold,anchor=south] at (axis cs:1,320) {312};
\node[font=\tiny,text=transferreuse,anchor=south] at (axis cs:2,126) {118};
\node[font=\tiny,text=transferverdi,anchor=south] at (axis cs:3,104) {96};
\end{axis}
\end{tikzpicture}\hfill
\begin{tikzpicture}[baseline=(current bounding box.north)]
\begin{axis}[transferpanel,title={(c) Coverage and transfer risk},
  xmin=.1,xmax=1.04,ymin=0,ymax=.40,
  xtick={.2,.6,1.0},xticklabels={0.2,0.6,1.0},ytick={0,.2,.4},
  xlabel={Accepted coverage},ylabel={Neg. transfer risk}]
\addplot[only marks,mark=*,mark size=3pt,transferverdi]
  coordinates {(.47,.06)};
\addplot[only marks,mark=square*,mark size=2.8pt,transferreuse]
  coordinates {(1,.34)};
\node[font=\tiny,text=transferverdi,anchor=south]
  at (axis cs:.47,.085) {VERDI};
\node[font=\tiny,text=transferreuse,anchor=east]
  at (axis cs:.97,.34) {Unconditional};
\end{axis}
\end{tikzpicture}
\endgroup
  \caption{\textbf{Transfer versus cold start and unconditional reuse.}
  (a,b) Mean trials and measured GPU-h across 27 campaigns on nine backbones.
  (c) Coverage and conditional negative-transfer risk across 216 decisions.
  All panels show point estimates.}
  \label{fig:lobo-summary}
  
\end{figure}

\subsection{Comparison with Coding Agents and Human Experts}
\label{sec:exp-external}

Table~\ref{tab:external-comparison} compares coding agents and human experts
across nine backbones and 27 campaigns, with a shared registry, verifier, and
200 GPU-h nominal allowance. VERDI completes 24 campaigns, versus 20 for
Codex and 19 each for Claude Code and experts, and requires an average of 3.6 candidate trials to reach
the first verifier-confirmed repair, compared with 4.6 candidate trials for Codex, 4.9 for Claude Code, and 7.4 for human experts. VERDI, Codex, and
Claude Code attain comparable final utility, while human experts attain higher
final utility. Native executors differ, so this comparison measures end-to-end
performance rather than isolating the control scaffold.
Matched-GPT-5.5 controls are in App.~\ref{app:agent-controls}.

\begin{table}[!htb]
  \setlength{\abovecaptionskip}{0pt}
  \setlength{\belowcaptionskip}{0pt}
  \centering
  \begin{minipage}[t]{0.49\linewidth}
    \centering\vspace{0pt}
    \caption{\textbf{Selection-signal ablation.} 216 held-out decisions;
    27 LOBO campaign--seed units.}
    \label{tab:selector-main}
    \setlength{\tabcolsep}{2pt}
    \begin{tabular}{@{}lrrrr@{}}
      \toprule
      Selector & \makecell{Sign\\acc.$\uparrow$} &
      \makecell{Trials\\$\downarrow$} & \makecell{Risk\\$\downarrow$} &
      \makecell{Viol.\\$\downarrow$} \\
      \midrule
      Random & .50 & 8.7 & .38 & .33 \\
      Metadata & .65 & 5.1 & .21 & .13 \\
      Raw response & .71 & 4.4 & .15 & .09 \\
      Learned emb. & .74 & 4.6 & .17 & .10 \\
      Full fingerprint & \textbf{.83} & \textbf{3.6} & \textbf{.06} & \textbf{.03} \\
      \bottomrule
    \end{tabular}
  \end{minipage}\hfill
  \begin{minipage}[t]{0.49\linewidth}
    \centering\vspace{0pt}
    \caption{\textbf{External systems.} Completion and trials over 27 campaigns.}
    \label{tab:external-comparison}
    \setlength{\tabcolsep}{2pt}
    \begin{tabular}{@{}lcc@{}}
      \toprule
      System & \makecell{Done\\$\uparrow$} & \makecell{Trials\\$\downarrow$} \\
      \midrule
      VERDI (GPT-5.5) & \textbf{24/27} & \textbf{3.6} \\
      Codex (GPT-6) & 20/27 & 4.6 \\
      Claude Code (Fable-5) & 19/27 & 4.9 \\
      Human experts & 19/27 & 7.4 \\
      \bottomrule
    \end{tabular}
  \end{minipage}
  
\end{table}

\subsection{Ablation Studies}
\label{sec:exp-ablations}
\label{sec:exp-probe-evolution}

We ablate four decisions in VERDI: representation, control scaffold, probe
admission, and protected-outcome checks. These tests separate candidate
ranking from permissive acceptance.

\paragraph{Selection representation.}
Table~\ref{tab:selector-main} varies only the selector with the executor, archive,
Operator registry, and verifier fixed. Full fingerprint reaches 0.83 effect-sign
accuracy on 216 decisions, versus 0.71 for raw responses and 0.74 for a learned
embedding, while also reducing trials, negative transfer, and protected violations.

\paragraph{Control scaffold under a fixed executor.}
The external comparison also changes the executor. With GPT-5.5 fixed, removing
eligibility checks raises trials from 3.6 to 4.3 and risk from 0.06 to 0.21,
while removing IRG gives 4.8 trials and 0.18 risk (Table~\ref{tab:executor-control}).
Bare and generic loops require 187--213 GPU-h versus 96 for the full system.

\begin{table}[!htb]
  
  \setlength{\abovecaptionskip}{0pt}
  \setlength{\belowcaptionskip}{0pt}
  \centering
  \caption{\textbf{Control-scaffold ablation.}
  Nine backbones and 27 campaigns with the same Operator registry, frozen verifier,
  paired contexts, and nominal 200-GPU-h allowance. $C_{\mathrm{first+}}$ is
  measured GPU-h to the first verified positive; lower is better.}
  \label{tab:executor-control}
  \small
  \setlength{\tabcolsep}{4pt}
  \begin{tabular}{@{}lrrrr@{}}
    \toprule
    Configuration & Trials $\downarrow$ & $C_{\mathrm{first+}}$ (GPU-h) $\downarrow$ &
    Neg. transfer $\downarrow$ & Prot. viol. $\downarrow$ \\
    \midrule
    w/o eligibility checks & 4.3 & 128 & .21 & .14 \\
    w/o IRG & 4.8 & 142 & .18 & .11 \\
    Bare LLM agent & 5.9 & 187 & .31 & .26 \\
    Generic research-agent loop & 6.4 & 213 & .36 & .29 \\
    Full VERDI & \textbf{3.6} & \textbf{96} & \textbf{.06} & \textbf{.03} \\
    \bottomrule
  \end{tabular}
  
\end{table}

\paragraph{Probe refinement.}
Starting from four probes, gated growth admits three more, reducing nested
regret from 0.29 to 0.21 and calibration error from 0.14 to 0.08; unrestricted
growth uses fourteen probes but worsens both to 0.31 and 0.19
(Figure~\ref{fig:atlas-evolution}).

% Keep the final experiment figure before the concluding text.
\begin{figure}[H]
  \setlength{\abovecaptionskip}{0pt}
  \setlength{\belowcaptionskip}{0pt}
  
  \centering
  \begingroup
\definecolor{atlasfixed}{HTML}{66758B}
\definecolor{atlasfree}{HTML}{C68B53}
\definecolor{atlasgated}{HTML}{287D8E}
\pgfplotsset{atlaspanel/.style={
  width=.305\linewidth,height=2.9cm,
  ybar,bar shift=0pt,bar width=13pt,xmin=.45,xmax=3.55,ymin=0,
  xtick={1,2,3},xticklabels={Fixed,Free,Gated},
  tick label style={font=\scriptsize},title style={font=\small},
  axis x line*=bottom,axis y line*=left,
  axis line style={black!35},tick style={black!35},
  ymajorgrids,grid style={black!9},
  nodes near coords,nodes near coords style={font=\scriptsize},
  every node near coord/.append style={/pgf/number format/fixed,
    /pgf/number format/precision=2},
  enlarge x limits=false,clip=false}}
\begin{tikzpicture}
\begin{axis}[atlaspanel,title={(a) Selection regret $\downarrow$},
  ymax=.39,ytick={0,.1,.2,.3}]
\addplot[fill=atlasfixed,draw=none] coordinates {(1,.29)};
\addplot[fill=atlasfree,draw=none] coordinates {(2,.31)};
\addplot[fill=atlasgated,draw=none] coordinates {(3,.21)};
\end{axis}
\end{tikzpicture}\hfill
\begin{tikzpicture}
\begin{axis}[atlaspanel,title={(b) Calibration error $\downarrow$},
  ymax=.245,ytick={0,.1,.2}]
\addplot[fill=atlasfixed,draw=none] coordinates {(1,.14)};
\addplot[fill=atlasfree,draw=none] coordinates {(2,.19)};
\addplot[fill=atlasgated,draw=none] coordinates {(3,.08)};
\end{axis}
\end{tikzpicture}\hfill
\begin{tikzpicture}
\begin{axis}[atlaspanel,title={(c) Probe count},
  ymax=18,ytick={0,5,10,15}]
\addplot[fill=atlasfixed,draw=none] coordinates {(1,4)};
\addplot[fill=atlasfree,draw=none] coordinates {(2,14)};
\addplot[fill=atlasgated,draw=none] coordinates {(3,7)};
\end{axis}
\end{tikzpicture}
\endgroup
  \caption{\textbf{Probe-growth ablation.}
  Fixed: four probes; Free: unrestricted growth; Gated: admission from opposing
  outcomes. Regret is relative to the best admissible operator.}
  \label{fig:atlas-evolution}
  
\end{figure}

\paragraph{Protected-outcome verification.}
Removing negative-correlation gates raises nominal gain from +12.78 to +14.92
but also raises protected violations from 0.03 to 0.41 (App.~\ref{app:gate});
a larger scalar gain cannot substitute for a verified improvement.

\section{Conclusion}
\label{sec:conclusion}

We studied how optimization experience can accumulate across world models that look alike yet can
require opposite interventions. \methodname{} treats verified disagreement as evidence: it probes
each backbone, retrieves operators through the Interventional Repair Geometry, and settles paired
target experiments under a frozen verifier; verified contradictions then repair the description
used by later campaigns. Across nine backbones, this loop predicts effect signs with 83\%
accuracy.
It reduces trials to a first verified improvement by 68\% relative to cold start and achieves 0.06 negative-transfer risk at 47\% coverage. Future work should re-measure fingerprints and local effects after material changes to architecture, data, or hooks, and give each new objective its own verifier version.
\section*{AI Use Statement}
\label{sec:ai-use}

We used generative AI tools in two roles. First, GPT-5.5 is the explicitly
evaluated executor in VERDI: under a frozen prompt and tool schema, it proposes
candidate operators, implements them through backbone adapters, and executes the
declared search and evaluation workflow. Second, language models assisted with
manuscript editing, LaTeX diagnostics, figure and table layout. The authors retain responsibility for
validating AI-assisted text, code, figures, and scientific claims and for the
final content of this work.

% ------------------------------------------------------------
\section*{Ethics Statement}
\label{sec:ethics}

VERDI improves auditability but does not guarantee truth; results require
independent replication, human review, and disclosure of generative-agent use.
Training jobs, repository edits, and robot access require isolated credentials,
least privilege, sandboxing, human approval, and an independent safety layer.
Verify licenses and privacy for robot data/checkpoints, record provenance without
secrets, assess dual-use risks, and use staged evaluation with explicit compute
accounting.

% ------------------------------------------------------------
\section*{Reproducibility Statement}
\label{sec:repro}

Baseline and operator-applied arms share initialization, steps, data, evaluator, paired
contexts/seeds, target hiding, archive visibility, Operator registry, and nominal
allowance; measured costs are separate. The GPT-5.5 executor runs with a frozen
prompt and tool schema. Protocols and controlled comparisons are detailed in
Appendix~\ref{app:ablations}; base models and datasets are publicly available.

% ------------------------------------------------------------
\bibliographystyle{iclr2027_conference}
\bibliography{iclr2027_conference}
\clearpage
\appendix

\section{Formal Method Details}
\label{app:system}

We study automated improvement of a pretrained world model under a user-specified goal and a finite verification allowance. Small, controlled inference-time probes provide paired response Jacobians, which describe how distinguishable behavioral changes appear along different probe directions. \methodname{} organizes these responses into an Interventional Repair Geometry (IRG) and uses compatible local charts to rank operator outcome evidence across backbones. When nearby models show opposite effects for the same operator, that disagreement supplies the next diagnostic direction and updates the geometry for later campaigns.

\subsection{Goal-Conditioned World-Model Improvement}
\label{app:problem}

A campaign is a tuple $\cC=(b,\theta_0,\mathcal{D},G,V_G,\cB)$, where $b$ is a backbone with pretrained parameters $\theta_0$, $\mathcal{D}$ is the campaign data, $G$ is a goal specification, $V_G$ is an independent verifier, and $\cB$ is a nominal resource allowance. The goal specifies a primary utility $U_G$, evaluation contexts $x\sim\mathcal{D}^{\mathrm{ver}}$, horizons $h\in\mathcal{H}_G$, and a set of validity constraints $v_j$. For a candidate model $\theta$, we define
\begin{equation}
J_G(\theta)=\sum_{h\in\mathcal{H}_G}w_h\,
\E_{x\sim\mathcal{D}^{\mathrm{ver}}}
[U_G(\theta;x,h)],
\quad\text{s.t.}\quad v_j(\theta)\leq 0\ \ \forall j.
\label{eq:goal}
\end{equation}
The constraints make the utility contract explicit: an operator must improve the target while preserving action use, motion, and other declared behaviors. Our outcome schema follows the multi-dimensional world-model evaluation suite used in the experiments, with tolerances calibrated from the observed relationships among metrics. All comparisons use paired evaluation contexts and seeds. For an operator set $S$, we retain $\bm{y}(S)=[\Delta J_G(S),\{\Delta U_k(S,h)\}_{k,h},\{v_j(S)\}_j,C(S)]$, where $C(S)$ is measured cost. Keeping this vector makes it possible to follow an operator across short-horizon fidelity, long-horizon dynamics, and behavioral validity instead of reducing the campaign to one score.

\subsection{Interventional Repair Geometry}
\label{app:irg}

A static probe bank first extracts $\bm{z}_x=P_v(\theta,\mathcal{D}^{\mathrm{dev}},G)$ for campaign instance $x$. These inexpensive observations condition effect prediction; the geometric coordinates come from paired intervention responses so that their meaning is shared across backbones.

The backbone adapter declares a capability vector $\bm{c}_x$ and a set of typed intervention paths $\mathcal{I}_{b,s}(d)$ with semantic dose $d$ and $\mathcal{I}_{b,s}(0)=\mathrm{Id}$. A path must be inference-only, reversible, paired with the same seeds and trajectories as the baseline, and measurable at at least two nonzero doses. Examples include action scaling, controlled context retention, first-frame anchoring strength, and sampler-noise stress. Training losses and data reweighting are excluded because their optimization response is neither cheap nor dose-comparable across backbones.

Let $\bm{q}_x(\bm{d})$ denote the robustly standardized vector of goal, diagnostic, and validity outcomes under intervention dose vector $\bm{d}$. In chart $q$, paired central differences estimate the local response Jacobian
\begin{equation}
\bm{J}_x^{q}[:,s]=
\left.\frac{\partial \E[\bm{q}_x(\bm{d})]}{\partial d_s}\right|_{\bm{d}=0}
\approx
\frac{\widehat{\bm{q}}_x(+\delta_s)-
      \widehat{\bm{q}}_x(-\delta_s)}{2\delta_s}.
\label{eq:response-jacobian-app}
\end{equation}
For a one-sided path, the compiler records a normalized secant instead. A small dose grid checks that the response is locally stable; paths whose finite-difference slope changes beyond tolerance remain finite-dose probes rather than local directions.

The Jacobian induces a weighted IRG metric on the intervention frame,
\begin{equation}
\bm{G}_x^{q}=(\bm{J}_x^{q})^\top\bm{W}_q\bm{J}_x^{q}+\lambda\bm{I},
\qquad
\bm{r}_x^{q}=\operatorname{vec}(\bm{W}_q^{1/2}\bm{J}_x^{q}),
\label{eq:pullback}
\end{equation}
where $\bm{W}_q$ is fixed by the goal and outcome schema before the test campaign. Bootstrap repeats provide uncertainty $\bm{\Sigma}_x^{q}$. A chart records the goal and outcome schema, capability class, and supported intervention frame; compatible charts form the IRG. Static probes $\bm{z}_x$, capabilities $\bm{c}_x$, and missingness metadata condition the effect predictor. Starting from a small canonical frame, we retain an additional direction only when nested cross-domain replay shows that it improves operator selection enough to justify its probing cost.

\subsection{Operator Descriptors and Compilation}
\label{app:typed}

An operator is represented by a descriptor $a=(\ell_a,\psi_a,\omega_a,d_a,s_a,\mathsf{pre}_a,\mathsf{inv}_a,\bm{\chi}_a)$, where $\ell_a$ is a required hook type, $\psi_a$ is the transformation, $\omega_a$ is its scope, $d_a$ is a behaviorally defined dose, $s_a$ is a schedule, $\mathsf{pre}_a$ lists applicability conditions, $\mathsf{inv}_a$ lists semantic invariants, and $\bm{\chi}_a$ is a falsifiable prediction over operator response and horizon behavior. Examples include preserving the data split, applying a loss to the declared tensor, or restricting a sampler change to inference. The Operator registry stores these descriptors; the Operator outcome archive records their confirmed positive, null, or harmful outcomes by task type and capability class.

Each backbone exposes a capability vector $\bm{c}_b$ and a typed hook inventory. A compiler maps $a$ to $\Gamma_b(a)=(\delta_a,r_a)$ when the preconditions and invariants are satisfied. The receipt records hook-contract tests, dose units and direction, outcome semantics, and invariant checks. For cross-backbone comparison, compatible receipts propose a partial intervention-axis map $\bm{T}_{i\rightarrow x}$ and outcome map $\bm{B}_{i\rightarrow x}$; the maps are then checked against paired response measurements.

\subsection{Capability-Preserving Geometric Transfer}
\label{app:transfer}

Compatible receipts define a shared chart and maps $\bm{T}_{i\rightarrow x}$ and $\bm{B}_{i\rightarrow x}$. We reject a map when paired anchor contexts do not preserve the measured response directions:
\begin{equation}
e_{\mathrm{align}}(i,x)=
\frac{\|\bm{J}_x\bm{T}_{i\rightarrow x}-
\bm{B}_{i\rightarrow x}\bm{J}_i\|_F}
{\|\bm{J}_x\bm{T}_{i\rightarrow x}\|_F+
 \|\bm{B}_{i\rightarrow x}\bm{J}_i\|_F+\epsilon}.
\label{eq:alignment}
\end{equation}
Within a fixed shared chart, a simple fixed, uncertainty-normalized distance defines IRG neighbors:
\begin{equation}
d_{\mathrm{IRG}}^2(x,i)=
(\bm{r}_x-\widetilde{\bm{r}}_i)^\top
\bm{W}_{q}^{\mathrm{dist}}(\bm{r}_x-\widetilde{\bm{r}}_i)
+\lambda_c d_{\mathrm{cap}}^2(\bm{c}_x,\bm{c}_i),
\label{eq:irg-distance}
\end{equation}
where tildes denote transported coordinates and $\bm{W}_{q}^{\mathrm{dist}}$ is fixed from the frozen, target-independent archive $M_k^{\mathrm{full}}$ before target evaluation, with noisy coordinates downweighted. The archive contains shared prior entries and all eligible pre-target source campaigns; target-specific records and post-evaluation updates are excluded. Each operator $a$ defines a local effect function $\tau_a$ over this space, estimated only from campaigns satisfying the same operator contract. Thus transfer predicts a context-local effect, not the universal validity of a recipe.

A weighted local predictor gives mean effect $\widehat\mu_a(x)$ and uncertainty $\widehat\sigma_a(x)$. The geometry version, predictor, and operator set are frozen before an independent group-calibration split constructs a one-sided lower bound $L_a(x)$. Campaigns, not frames, are calibration units, and a changed IRG version invalidates the old bound.

Transfer prioritization uses an eligibility check, while effect acceptance comes from
the subsequent target-side experiment. Protected violations $v_j\leq0$ are
measured after execution and are not guessed from source records. Formally,
\begin{equation}
\begin{split}
    \mathsf{Cert}_{\mathrm{app}}(x,a) ={}& \mathbb{I}[\mathsf{compile}=1]
    \mathbb{I}[\mathsf{overlap}\geq\kappa_{\min}]
    \mathbb{I}[N_{\mathrm{eff}}\geq n_{\min}]\\
    &\cdot \mathbb{I}[e_{\mathrm{align}}\leq\epsilon_{\mathrm{align}}]
    \mathbb{I}[\mathsf{sign\_agree}\geq\gamma]
    \mathbb{I}[L_a(x)>\delta_G],
    \label{eq:certificate-app}
\end{split}
\end{equation}
The six terms cover compilation, support overlap, evidence, response alignment,
sign agreement, and predicted gain. A failed term removes transfer priority for
that candidate. The system can still reuse lower levels of infrastructure: L0
transfers the loop, L1 transfers compatible diagnostics, L2 transfers a
compiled operator, and L3 transfers an effect prior under
Eq.~\ref{eq:certificate-app}. In every case, the target experiment settles the
effect.

Among valid candidates, the online selector is deliberately simple:
\begin{equation}
a_t=\arg\max_{a\in\mathcal{A}_t}
\frac{\widehat\mu_a(x)+\beta_t\widehat\sigma_a(x)}
{\widehat{C}(a)}.
\label{eq:acquisition}
\end{equation}
Validity and capability constraints define $\mathcal{A}_t$ and are not traded against reward through tunable penalties. A transfer that passes these checks initializes the posterior; an abstention uses a cold prior.

We use progressive fidelity: a low-cost paired screen estimates effect direction, and promoted candidates receive paired multi-seed confirmation on held-out contexts. Screening updates search uncertainty; only confirmation creates a settled effect record.

\subsection{Individual Operator Outcomes and Interactions}
\label{app:credit}

Operator combinations are useful for exploration but ambiguous for attribution. If a set $S$ is promising, \methodname{} schedules matched contrasts using common seeds and examples. For operator $i\in S$, its local contribution is estimated from the available factorial contrasts
\begin{equation}
\widehat{\tau}_i=
\sum_{A\in\mathcal{F}_i}w_A
\left[\overline{Y}(A\cup\{i\})-\overline{Y}(A)\right],
\label{eq:credit}
\end{equation}
where $\mathcal{F}_i\subseteq 2^{S\setminus\{i\}}$ contains matched contexts. With all subsets, Eq.~\ref{eq:credit} recovers Shapley-style factorial credit; under a smaller allowance, leave-one-out effects are retained explicitly as conditional rather than universal effects. Pairwise interaction is stored separately as $\widehat{\tau}_{ij}^{\mathrm{int}}=\overline{Y}(\{i,j\})-\overline{Y}(\{i\})-\overline{Y}(\{j\})+\overline{Y}(\varnothing)$.

An entry becomes a \emph{confirmed positive operator outcome} when its lower confidence bound exceeds the campaign threshold, all validity gates pass, protected metrics remain within tolerance, and the result replicates over the required seeds. Its context records the IRG chart, backbone capabilities, data regime, goal, and horizon range. The Operator outcome archive stores validated positive, null, harmful, and interaction outcomes, so later campaigns can use both successful and unsuccessful evidence.

\subsection{Probe Refinement}
\label{app:evolution}

A fixed probe basis can miss distinctions relevant to operator selection. We identify pairs of campaigns with the same goal schema and shared operator support that are close in IRG but have reliable, opposing outcomes for the same operator. Let
\begin{equation}
\begin{aligned}
\mathcal{K}_{\tau}=\Bigl\{(i,j): \;&d_{\mathrm{IRG}}(i,j)\leq\epsilon_{\mathrm{IRG}},\\
&\exists a\in\mathcal{A}_i\cap\mathcal{A}_j:\\
&\operatorname{LCB}(-\tau_{i,a}\tau_{j,a})>0,\\
&\min(|\tau_{i,a}|,|\tau_{j,a}|)>\epsilon_c\Bigr\}.
\end{aligned}
\label{eq:collision-app}
\end{equation}
Here $\tau_{i,a}$ denotes the utility effect $r_i^U(a)$ in the main text.
The set above formalizes the opposing outcomes described in the main text,
using a confidence bound on the negative effect product and a minimum-effect filter.
An operational sufficient check is that the individual effect intervals lie
on opposite sides of zero, each beyond $\epsilon_c$, together with the
sign-error screening criterion. A generic confidence bound is not assumed
to be a posterior probability.
Such a disagreement marks a distinction relevant to operator selection that the current chart has missed. In our audits of opposing outcomes, this can arise when a feature-conditioning operator improves soft-body scenes but degrades several metrics on rigid-body scenes whose static descriptions look similar. The system requests a new diagnostic direction when the disagreement survives uncertainty and schema checks. From candidate probes, it selects
\begin{equation}
s^\star=\arg\max_s
\frac{\LCB\!\left[
\widehat{\mathcal{R}}_{\mathrm{nested}}(\mathcal{S})-
\widehat{\mathcal{R}}_{\mathrm{nested}}(\mathcal{S}\cup\{s\})
\mid\mathcal{K}_{\tau}\right]}
{C_{\mathrm{probe}}(s)},
\label{eq:collision-separation}
\end{equation}
which directly targets future operator-selection errors rather than raw feature separation. Because disagreements are screened over many campaign pairs, we control the false discovery rate over $\mathcal{K}_{\tau}$ and shrink small-sample effects toward the chart mean before staging a direction. The inner replay learns the frame and effect predictor; disjoint promotion domains measure regret and calibration. Repeated search failure without such a disagreement instead points to a missing operator and prompts search for additional operators.

The executor searches network-accessible literature, documentation, and
repositories for cross-domain ideas and synthesizes target-adapter code. A new
candidate first runs in shadow mode under nested leave-one-domain/backbone-out
replay. It enters the next registry version when it lowers selection regret and
trials-to-first-positive without worsening calibration or the frozen regression
battery. Registry snapshots, verdict-facing probes, and $V_G$ stay fixed within
a campaign, so discovery can expand the library without changing an ongoing
evaluation.

\paragraph{What is learned.}
The result is an IRG together with verified operators, local outcomes and
interactions, transfer eligibility checks, negative results, and replay-tested
capabilities. Each settled experiment either improves the local effect model,
marks a transfer boundary, or exposes opposing outcomes that guide the next IRG
revision.

\begin{algorithm}[tbp]
\caption{IRG Learning and Selective Operator Transfer in \methodname{}}
\label{alg:verdiwm}
\begin{algorithmic}[1]
\Require Campaign $\cC$, probes $P_v$, intervention frame $\mathcal{S}_v$, Operator registry $\mathcal{R}_v$, Operator outcome archive $\mathcal{E}$
\Ensure Verified operator $A^\star$, updated $\mathcal{E}$, staged capabilities
\State Instantiate and freeze verifier $V_G$; audit backbone capabilities $\bm{c}_b$
\State Measure static covariates $\bm{z}$ and compile the supported intervention frame $\mathcal{S}_b\subseteq\mathcal{S}_v$
\State Run paired dose paths; estimate $(\bm{J}_x,\bm{G}_x,\bm{\Sigma}_x)$ and locality residuals
\State Query literature, documentation, and repositories with the frozen Optimization Fingerprint; type-check and shadow-stage candidates for $\mathcal{R}_{v+1}$
\For{each operator $a\in\mathcal{R}_v$}
  \State Compile $a$ and chart maps; reject failed contracts or excessive map distortion
  \State Evaluate $\mathsf{Cert}(x,a)$; initialize a transported-field or cold posterior
\EndFor
\State Let $\mathcal{A}$ be the remaining compiled candidates
\While{remaining nominal allowance is sufficient for a valid trial}
  \State Select $a$ from $\mathcal{A}$ using Eq.~\ref{eq:acquisition}
  \State Run the cheapest admissible paired screen; log response $\bm{y}(a)$ and receipts
  \State Update search uncertainty; keep screen evidence provisional
  \If{$a$ passes the low-cost promotion rule}
    \State Run held-out multi-seed confirmation under frozen $V_G$
    \If{$a$ is composite and remains positive}
      \State Schedule matched factorial or leave-one-out contrasts
    \EndIf
    \State Mark only replicated, gate-valid positive operator outcomes as confirmed
    \State Archive the verifier-settled operator outcome, including null or harmful outcomes, in $\mathcal{E}$
  \EndIf
  \If{nearby models have reliable opposing operator outcomes $\mathcal{K}_{\tau}$}
    \State Stage the direction maximizing nested-replay regret reduction in Eq.~\ref{eq:collision-separation}
  \ElsIf{search stagnates without such a disagreement}
    \State Stage a new operator for later campaigns
  \EndIf
\EndWhile
\State Shadow-replay staged capabilities on disjoint promotion domains; queue only regret- and calibration-valid passes for version $v+1$
\State \Return best verified operator $A^\star$, $\mathcal{E}$, and the staged queue
\end{algorithmic}
\end{algorithm}

% Supplementary ablations and audit details.

\section{Ablation and Audit Details}
\label{app:ablations}

The complete LOBO design covers nine target backbones with three seeds each
(27 campaigns) and 216 source--target--operator decisions
(Table~\ref{tab:shared-prior-archive}). The detailed selector, archive-replay,
and fixed-executor point estimates reported below use the three-backbone
subset with nine campaigns and 72 decisions; the nine-target table records the
full study inventory. Each campaign pairs
baseline and operator-applied models on matched contexts under the frozen verifier $V_G$; trials and cost
are averaged over the 27 backbone--seed campaigns. Coverage is the fraction of
hypotheses that pass the transfer-prioritization gate, and negative-transfer
risk is measured among the accepted transfers that are executed.

Archive-prefix replay, selector ablations, and fixed-executor comparisons use
the detailed subset unless a table explicitly states the full nine-target
scope. Confidence intervals use backbone-cluster bootstrap with all seeds
retained within each sampled backbone. Additional testbeds and audits state
their scope separately.

\paragraph{Archive ablation convention.}
For held-out target $t$ in fold $k$, the complete archive is denoted $M_k^{\mathrm{full}}$ and is frozen before target-side evaluation. It contains the shared prior archive and all eligible pre-target source campaigns; target-specific records, target-side calibration, tuning decisions, and post-evaluation updates are excluded. The archive ablation compares $M_k^{\mathrm{full}}$ against $M_k^{\mathrm{empty}}=\varnothing$ (the cold-start arm). All other ablations keep $M_k^{\mathrm{full}}$ fixed and vary only the named component; source and target protocols, budgets, and verifiers remain unchanged.

\paragraph{Eight-environment testbed (ACWM-Phys).}
The fingerprint and distance analyses use an eight-environment testbed built on
ACWM-Phys, a compact action-conditioned world model, spanning
rigid-body contact (push/stack cube), deformable and granular media (cloth,
rope, sand), fluid transport (pour water), and articulated dynamics (robot arm,
reacher). Its size makes paired multi-seed confirmation, the exhaustive
reference evaluation of \S\ref{app:screening}, and the planted-fault audits of
\S\ref{app:control-plane-ablation} feasible at full fidelity. The shared prior
archive also contains settled campaigns from other backbones; together with
eligible source campaigns, these entries constitute $M_k^{\mathrm{full}}$.
Table~\ref{tab:shared-prior-archive} lists all nine LOBO targets separately
from the ACWM-Phys seed testbed. Each target can contribute source evidence in
other folds, but its own records are excluded when it is held out. For held-out transfer, the archive snapshot and selector calibration are frozen before target-side evaluation, and target results are added only after the fold is settled.
Cross-backbone behavior of the geometry is evaluated separately in the main
campaigns (\S\ref{sec:experiments}).

\begin{table}[t]
  \centering
  \caption{\textbf{Backbone coverage of the LOBO study.} All nine named
  backbones serve as held-out targets, totaling 27 campaigns and 216
  transfer decisions; source records exclude the target in each fold.
  All nine also enter archive replay and external-system comparisons.
  ACWM-Phys supplies a separate seed and diagnostic testbed.}
  \label{tab:shared-prior-archive}
  \small
  \begin{tabularx}{\linewidth}{@{}lX@{}}
    \toprule
    Backbone / testbed & Evaluation role \\
    \midrule
    Ctrl-World & LOBO, archive replay, and external comparison \\
    Cosmos-Predict~2 & LOBO, archive replay, and external comparison \\
    Cosmos3 & LOBO, archive replay, and external comparison \\
    IRASim & LOBO, archive replay, and external comparison \\
    OccWorld & LOBO, archive replay, and external comparison \\
    DreamerV3 & LOBO, archive replay, and external comparison \\
    TD-MPC2 & LOBO, archive replay, and external comparison \\
    DINO-WM & LOBO, archive replay, and external comparison \\
    V-JEPA~2 & LOBO, archive replay, and external comparison \\
    \midrule
    ACWM-Phys & Eight-environment seed and diagnostic testbed;
      outside the nine-target count \\
    \bottomrule
  \end{tabularx}
\end{table}

\subsection{Backbone identities and evaluation contracts}
\label{app:backbone-contracts}

The nine LOBO targets are distinct model interfaces. Every target uses the
same contract layers: a frozen goal and verifier, paired baseline/operator-applied
contexts, shared probe semantics translated by its adapter, and a target-specific
set of protected outcomes. Ctrl-World, Cosmos-Predict~2, and Cosmos3 additionally
have standalone case studies; IRASim, OccWorld, DreamerV3, TD-MPC2, DINO-WM,
and V-JEPA~2 contribute to the full LOBO transfer analysis.

\begin{table}[t]
\centering
\caption{\textbf{Identity and role of each LOBO target.} The ``role'' column
describes the model interface used by VERDI; the final column gives the
reported evaluation scope.}
\label{tab:backbone-contracts}
\small
\setlength{\tabcolsep}{3pt}
\begin{tabularx}{\linewidth}{@{}l>{\raggedright\arraybackslash}p{.25\linewidth}>{\raggedright\arraybackslash}X@{}}
\toprule
Backbone & Role in the evaluation & Paper evidence and contract scope \\
\midrule
Ctrl-World & Action-conditioned visual world model for robot manipulation &
LOBO target, archive replay, and model-specific case study. The case study
uses action/context/sampling probes, visual-quality coordinates, and protected
dynamic-behavior checks. \\
Cosmos-Predict~2 & Video/world-model predictor with representation and temporal prediction hooks &
LOBO target, archive replay, and model-specific representation-guidance case
study. The case study reports visual and structural coordinates together with
the declared protected checks. \\
Cosmos3 & World-action model for long-horizon simulation and control &
LOBO target, archive replay, model-specific simulation study, and real-robot
study. The reported contract includes rollout success, action-level diagnostics,
and protected horizon/control checks. \\
IRASim & Action-conditioned manipulation world model &
LOBO target; action--observation rollout contract and aggregate transfer
analysis. \\
OccWorld & Occupancy-oriented world model interface &
LOBO target; occupancy/free-space contract and aggregate transfer analysis. \\
DreamerV3 & Latent-imagination control model &
LOBO target; latent rollout/control contract and aggregate transfer analysis. \\
TD-MPC2 & Latent-dynamics model for model-predictive control &
LOBO target; latent dynamics and MPC-rollout contract and aggregate transfer
analysis. \\
DINO-WM & Visual-representation world model interface &
LOBO target; feature-space prediction/planning contract and aggregate transfer
analysis. \\
V-JEPA~2 & Video-prediction and representation model interface &
LOBO target; temporal representation/prediction contract and aggregate transfer
analysis. \\
\bottomrule
\end{tabularx}
\end{table}

Operationally, the targets span three interface groups. Ctrl-World,
Cosmos-Predict~2, and Cosmos3 are the detailed visual or world-action studies:
the first conditions visual prediction on robot actions, the second predicts
future video observations with representation guidance, and the third couples
visual prediction with action chunks for simulation and control. IRASim is an
interactive action-conditioned simulator whose contract is centered on
action--observation rollouts. OccWorld predicts future scene occupancy from
scene context and motion, so its contract uses occupancy and free-space
consistency. DreamerV3 and TD-MPC2 are latent-dynamics control interfaces:
their adapters expose observation/action histories, latent prediction, and
control rollout evaluation. DINO-WM predicts and plans in pretrained visual
feature space, while V-JEPA~2 provides a video representation/prediction
interface whose shared measurements focus on temporal and action-conditioned
response. The goal, utility scale, and protected thresholds are instantiated
per target contract; the shared layer specifies the measurements used for
cross-target transfer.

For each row, the adapter exposes the supported subset of action scaling,
context retention, first-frame anchoring, and sampler-noise probes. A probe is
retained only when its paired response has the required dose semantics and
measurement support. Operator hooks are recorded separately from probe hooks:
training, representation, sampling, and inference changes are compiled only
when the target adapter satisfies the declared preconditions. Thus the shared
contract defines what is compared, while the adapter defines how the same
measurement is executed on a particular backbone. Per-target trial counts,
protected outcomes, and fold-level decisions are reported as the audit units
for the nine-target aggregate; the three detailed case studies add
model-specific metrics and implementation traces.

\paragraph{Questions and evaluation quantities across targets.}
The nine-target design asks whether response-informed selection remains useful
across different model interfaces, whether it reduces target-side trials and
compute, and whether it controls negative transfer under each frozen goal
contract. Effect-sign accuracy compares predicted and observed signs of operator outcomes.
Trials and $C_{\mathrm{first+}}$ measure search to a verifier-confirmed
positive; coverage is the accepted fraction of candidate transfers, and risk
is the negative fraction among executed accepted transfers. The 216 decisions
are repeated observations within the 27 campaigns. Raw utilities from
different target tasks are reported within their own contracts. ACWM-Phys
diagnostics and RoboCoin adaptation are separate studies from the nine-target
LOBO count.

% ----------------------------------------------------------------------
\subsection{Additional Fingerprint Diagnostics}
The measured probe-response matrix and pairwise geometry below characterize the
underlying representation; selector comparisons are reported in
Table~\ref{tab:selector-main}.

\begin{figure}[t]
\centering
\includegraphics[width=\linewidth]{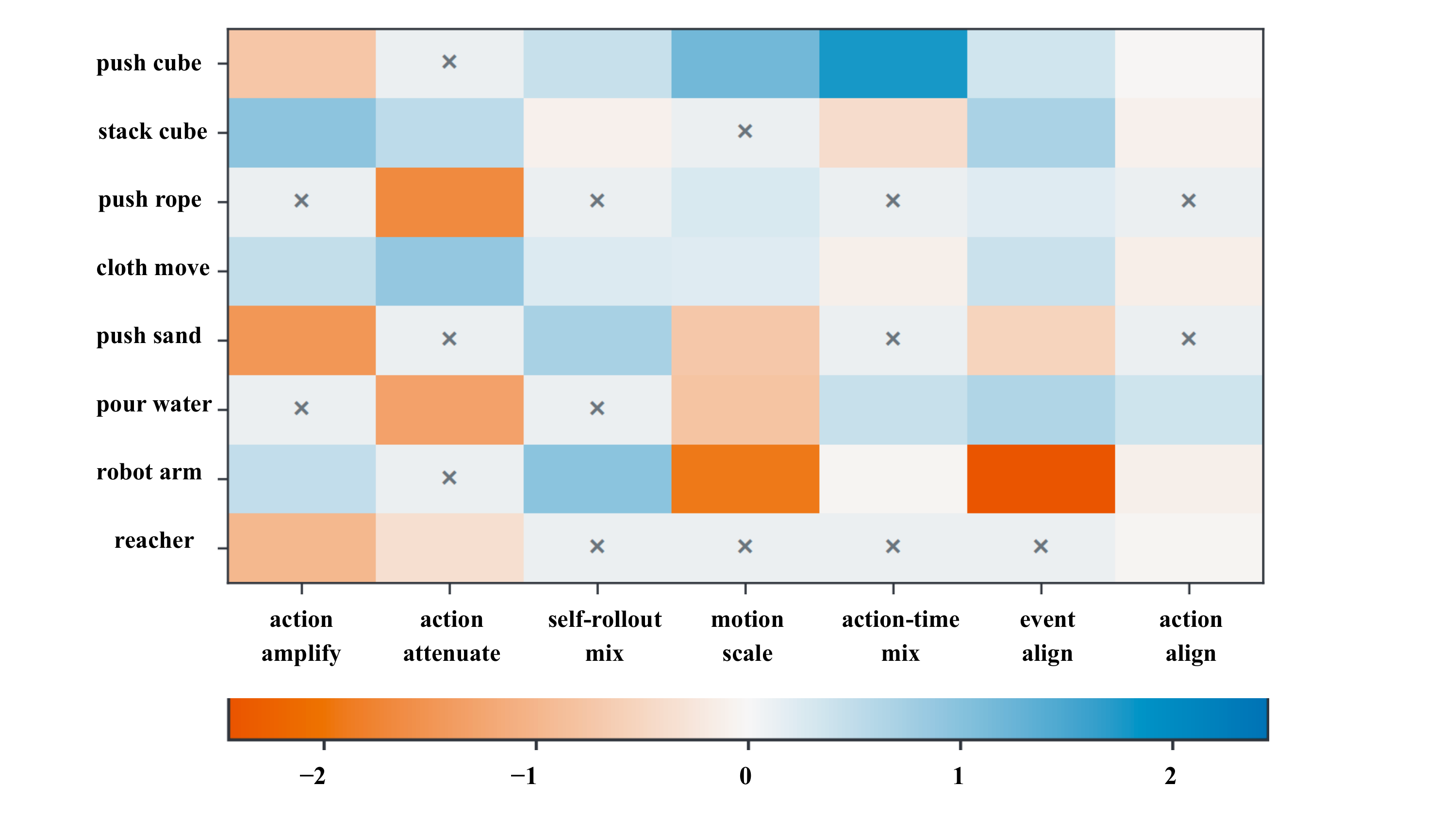}
\caption{\textbf{Dose-response slopes across eight environments and seven probe directions (ACWM-Phys testbed).}
All axes and the color scale are described here rather than in the panel.
Each cell is the paired finite-difference slope of the primary utility $U_G$ per unit dose---same seeds and trajectories, with and without a small dose---robustly standardized (\S\ref{app:irg}); rows (top to bottom) are the eight environments (push cube, stack cube, push rope, cloth move, push sand, pour water, robot arm, reacher), columns (left to right) the probe directions (action amplify, action attenuate, self-rollout mix, motion scale, action-time mix, event align, action align). Blue marks a positive slope, orange a negative one; saturation encodes magnitude ($|z|\approx 2$: strong response; $|z|\approx 0$: the path does not move $U_G$). Paths outside the supported intervention frame are marked $\times$. Columns 1--2 are the two one-sided directions of the canonical action-scaling path, recorded as normalized secants (\S\ref{app:irg}); columns 3--4 complete the canonical frame; columns 5--7 are diagnostic directions admitted in IRG v2 through probe refinement based on opposing operator outcomes, whose ablation is reported in main-text Figure~\ref{fig:atlas-evolution}.}
\label{fig:irg-atlas}
\end{figure}

% ----------------------------------------------------------------------
\begin{figure}[t]
\centering
\includegraphics[width=0.72\linewidth]{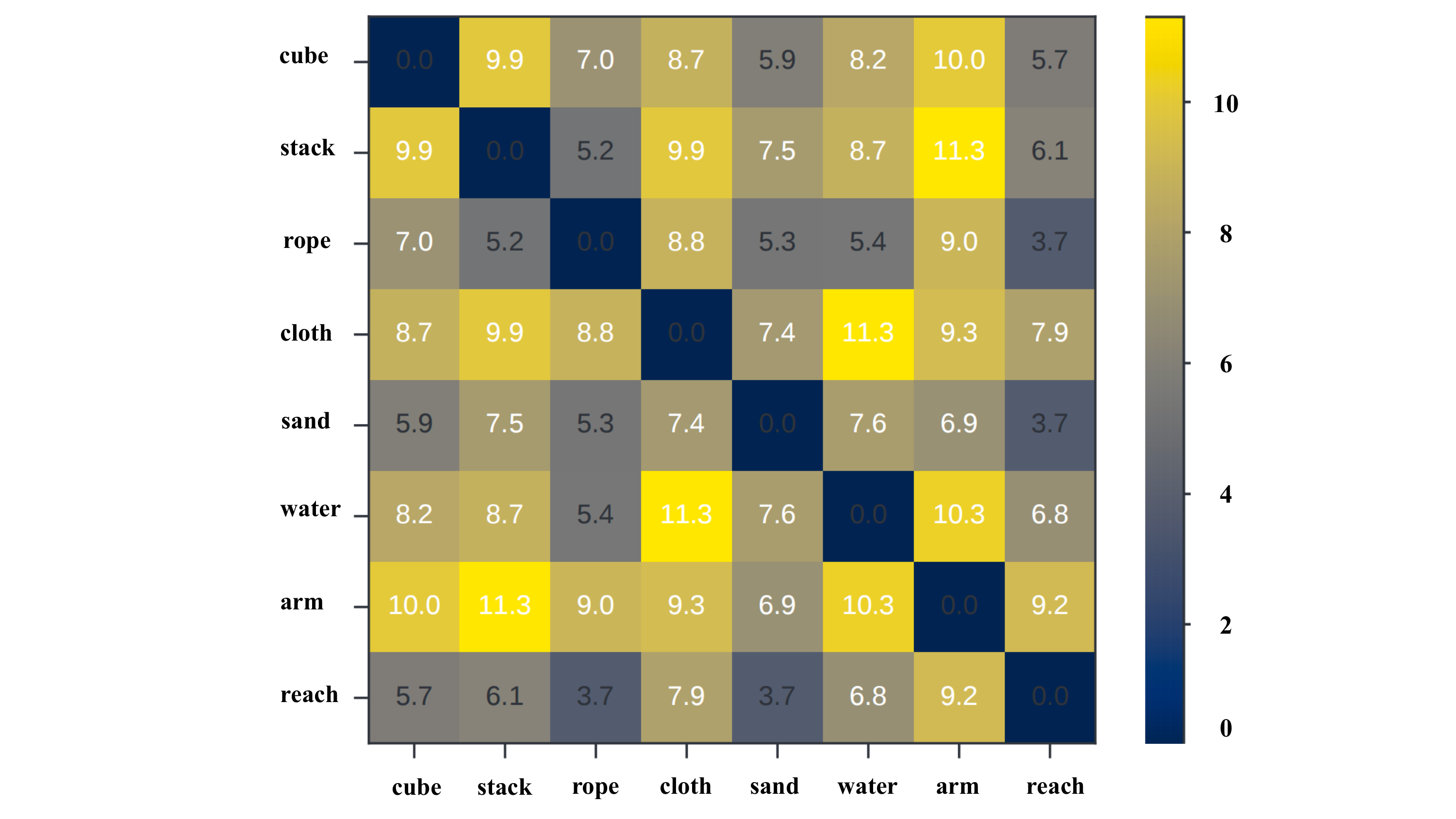}
\caption{\textbf{Pairwise IRG distances between environments (ACWM-Phys testbed).}
Cells show $d_{\mathrm{IRG}}$ from the full Optimization Fingerprint (Eq.~\ref{eq:irg-distance});
row/column order matches Fig.~\ref{fig:irg-atlas}, and the neighbor threshold is
$7.5$. The geometry places sand and rope within range ($d=5.3$) while separating
the superficially similar cube and stack tasks ($d=9.9$), defining the neighbors
used for local operator-effect estimation (\S\ref{app:transfer}).}
\label{fig:irg-distance}
\end{figure}

% ----------------------------------------------------------------------
\subsection{Agent-Level Controls Under Matched Budget}
\label{app:agent-controls}

\paragraph{Matched agent controls.}
The component ablations are complemented by matched agent controls. All arms
share the same frozen LLM, Operator registry snapshot, paired evaluation
protocol, seeds, and nominal per-campaign allowance ($\cB=200$ GPU-h; measured
overruns are retained). They differ only in the machinery used for proposal,
licensing, and settlement. The bare agent proposes and executes operators from
the registry, while the generic research-agent loop follows a
propose--implement--run--observe cycle. The frequency prior and random selector
provide simple references for historical frequency and uninformed choice.
End-to-end metrics follow main-text Figure~\ref{fig:lobo-summary}.

\begin{table}[ht]
\caption{Agent-level controls on the detailed three-backbone subset under a
matched nominal allowance (200 GPU-h per campaign; same LLM and Operator registry).
The full-system and random-selector results are reported in Table~\ref{tab:selector-main}
and Figure~\ref{fig:lobo-summary}.}
\label{tab:agent-controls}
\centering\fontsize{9}{10}\selectfont
\setlength{\tabcolsep}{4pt}
\begin{tabular}{lcccc}
\toprule
 & Trials-to- & $C_{\mathrm{first+}}$ & Neg.\ & Prot.\ \\
Agent configuration & first-pos.\ $\downarrow$ & (GPU-h) $\downarrow$ & transfer $\downarrow$ & viol.\ $\downarrow$ \\
\midrule
\quad w/o eligibility checks (IRG only) & 4.3 & 128 & 0.21 & 0.14 \\
\quad w/o IRG (cert.\ over text-sim.\ neighbors) & 4.8 & 142 & 0.18 & 0.11 \\
Bare LLM agent (same pool, free proposal) & 5.9 & 187 & 0.31 & 0.26 \\
Generic research-agent loop & 6.4 & 213 & 0.36 & 0.29 \\
Frequency prior (most-confirmed operator) & 5.2 & 164 & 0.29 & 0.22 \\
\bottomrule
\end{tabular}
\end{table}

Table~\ref{tab:agent-controls} complements the native-stack comparison in
Table~\ref{tab:external-comparison}: it holds the executor fixed and changes
only the search scaffold. Removing either IRG or eligibility checks increases failed
transfer and protected violations. The frequency prior tests whether a single
historically successful operator can substitute for target-specific selection.

\paragraph{Separate expert transfer case study.}
The human-expert row in Table~\ref{tab:external-comparison} aggregates actual
runs on the same 27 backbone--seed campaign units as the system comparison.
The small case study below is reported separately.
As an external reference, two engineers experienced with video world models
were given the same registry, nominal allowance, and verifier on Ctrl-World.
They reached first-positive in 3.9 trials with protected violation $0.05$---
comparable to \methodname{}---but required approximately 14 hours of expert
wall-clock per campaign and did not produce transferable records. Repeating the
exercise on Cosmos-Predict~2 reset their search nearly from scratch (10.8
trials), whereas VERDI's certified transfer required 3.9 trials. The comparison
points to the value of accumulating transferable evidence across campaigns,
rather than a difference in single-campaign intuition.

\subsection{Per-Backbone Transfer Breakdown}
\label{app:coverage-risk}

Table~\ref{tab:lobo-per-backbone} details transfer on Ctrl-World,
Cosmos-Predict~2, and Cosmos3 under the same protocol as the nine-backbone
aggregate in Figure~\ref{fig:lobo-summary}. It illustrates target-specific
variation within this subset.

\begin{table}[H]
\centering
\caption{\textbf{Detailed LOBO results for Ctrl-World and Cosmos backbones.} Each target is hidden while the
source archive and selector are frozen. Entries average three seeds. Coverage
and negative-transfer risk use the definitions in \S\ref{sec:exp-transfer}.}
\label{tab:lobo-per-backbone}
\fontsize{8}{9}\selectfont
\setlength{\tabcolsep}{3pt}
\begin{tabular}{@{}llcccc@{}}
\toprule
Held-out target & Transfer mode & \makecell{Trials to\\positive $\downarrow$} &
\makecell{$C_{\mathrm{first+}}$\\(GPU-h) $\downarrow$} & Coverage &
\makecell{Neg. transfer\\risk $\downarrow$} \\
\midrule
Ctrl-World & Cold start & 10.1 & 286 & -- & -- \\
 & Unconditional reuse & 3.8 & 104 & 1.00 & 0.31 \\
 & Eligibility-gated reuse & \textbf{3.3} & \textbf{84} & 0.50 & \textbf{0.04} \\
\midrule
Cosmos-Predict~2 & Cold start & 12.4 & 344 & -- & -- \\
 & Unconditional reuse & 4.5 & 132 & 1.00 & 0.37 \\
 & Eligibility-gated reuse & \textbf{3.9} & \textbf{108} & 0.44 & \textbf{0.08} \\
\midrule
Cosmos3 & Cold start & 11.1 & 306 & -- & -- \\
 & Unconditional reuse & 4.0 & 118 & 1.00 & 0.34 \\
 & Eligibility-gated reuse & \textbf{3.6} & \textbf{96} & 0.47 & \textbf{0.06} \\
\bottomrule
\end{tabular}
\end{table}

\subsection{Auditable Cross-Backbone Transfer Trace}
\label{app:transfer-trace}

Table~\ref{tab:transfer-trace} follows one complete decision alongside the
aggregate score. The Cosmos3 target and all of its records are hidden until
the source archive, IRG version, thresholds, and operator ranking are frozen.

\begin{table}[H]
\centering
\caption{\textbf{One cross-backbone transfer trace.} The eligibility gate
prioritizes a target experiment; the target-side verifier alone settles the
effect and authorizes the archive update. The target outcome concerns the
combined recipe and does not isolate the contribution of self-forcing.}
\label{tab:transfer-trace}
\fontsize{9}{10}\selectfont
\begin{tabularx}{.96\textwidth}{@{}>{\raggedright\arraybackslash}p{.22\textwidth}X@{}}
\toprule
Stage & Recorded evidence and decision \\
\midrule
Held-out target & Cosmos3 long-horizon rollout campaign; target records hidden during retrieval. \\
Retrieved source & ACWM-Phys robot-arm campaign, IRG v5; confirmed the self-forcing component under the shared rollout-drift contract. \\
Fingerprint support & $d_{\mathrm{IRG}}=0.43$; context-support overlap $=0.74$; common horizon support through 24 rollout steps. \\
Ranked hypothesis & Self-forcing component, retrieved rank 1; the Cosmos3 target recipe combines it with RL, while imported evidence changes search order but supplies no target claim. \\
Six pre-trial checks & compile $=1$; overlap $.74\geq.60$; $N_{\mathrm{eff}}=6.1\geq5$; alignment error $.11\leq.15$; sign agreement $.83\geq.75$; predicted-effect LCB $+2.3>\delta_G=1.5$. \\
Target screen & Paired low-cost screen: $\Delta U_G=+5.1$; promoted to confirmation, not written to the archive. \\
Held-out confirmation & Combined RL + self-forcing target recipe: paired three-seed effect $\Delta U_G=+6.8$ with 95\% CI $[3.2,10.4]$; all horizon and protected-coordinate gates pass. \\
Frozen-verifier verdict & Confirmed positive local effect for the declared context and horizon range. \\
Continual update & Write effect, uncertainty, support, cost, receipt, and artifact lineage to the Operator outcome archive; update the source--target outcome evidence for later campaigns. \\
\bottomrule
\end{tabularx}
\end{table}

\begin{table}[H]
\centering\small
\caption{\textbf{Verifier ablation.}
Error rates and verification cost (GPU-h) across verification policies.}
\label{tab:verifier-ablation}
\begin{tabular}{lcccc}
\toprule
Policy & False pos. & Harmful admit. & Protected viol. & Cost \\
\midrule
Absolute score & .22 & .19 & .17 & 84 \\
Unpaired & .18 & .16 & .14 & 91 \\
Paired, no gates & .11 & .09 & .08 & 97 \\
Paired + frozen verifier & .05 & .03 & .02 & 108 \\
Full eligibility check & \textbf{.03} & \textbf{.01} & \textbf{.01} & 114 \\
\bottomrule
\end{tabular}
\end{table}

Table~\ref{tab:verifier-ablation} shows why settlement is separated from
proposal. Unpaired or ungated policies admit more false positives and harmful
effects; pairing reduces variance, while the frozen verifier and full
eligibility check control protected violations. The extra compute pays for these
checks.

% ----------------------------------------------------------------------
\subsection{Eligibility Check Leave-One-Term-Out}
\label{app:cert-loo}

\paragraph{Eligibility checks.}
The six eligibility check terms guard distinct failure modes. We disable one at a
time in the LOBO replay of Figure~\ref{fig:lobo-summary}: failed compilation,
insufficient support overlap, too little effective source evidence,
response misalignment, sign disagreement, or an inadequate predicted gain.
Post-trial protected-validity checks remain active throughout.

\begin{table}[ht]
\caption{Eligibility Check leave-one-term-out. Each row disables exactly one term of Eq.~\ref{eq:certificate-app}; the reported rate is the failure mode that term is designed to guard. The full-system reference is reported only in main-text Figure~\ref{fig:lobo-summary}.}
\label{tab:cert-loo}
\centering\fontsize{9}{10}\selectfont
\begin{tabular}{llc}
\toprule
Disabled term & Exposed failure mode & Failure rate $\uparrow$ \\
\midrule
$\I[\mathsf{compile}=1]$ & compilation failure & 0.19 \\
$\I[\mathsf{overlap}\geq\kappa_{\min}]$ & insufficient support overlap & 0.17 \\
$\I[N_{\mathrm{eff}}\geq n_{\min}]$ & insufficient evidence & 0.15 \\
$\I[e_{\mathrm{align}}\leq\epsilon_{\mathrm{align}}]$ & response misalignment & 0.28 \\
$\I[\mathsf{sign\_agree}\geq\gamma]$ & sign disagreement & 0.18 \\
$\I[L_a(x)>\delta_G]$ & insufficient predicted gain & 0.24 \\
\bottomrule
\end{tabular}
\end{table}

Table~\ref{tab:cert-loo} reports the failure rate exposed by each omitted
check. Response alignment and the predicted-effect bound have the largest rates
in this audit. Compilation and support checks also prevent wasted trials; a
failed eligibility check falls back to a cold prior and the campaign can continue.

% ----------------------------------------------------------------------
\subsection{Validity-Gate Ablation and Threshold Sensitivity}
\label{app:gate}

\paragraph{Validity-gate sensitivity.}
The validity gates protect correlated partner metrics while the primary metric
improves. We compare the full gate set with a version that removes the
negative-correlation checks, and re-calibrate every tolerance $\tau$ at
$0.5\times$ and $2.0\times$ its default ($1.5\times$ the baseline seed-level
standard deviation).

\begin{table}[ht]
\caption{Validity-gate ablation and threshold sensitivity. Primary $\Delta$ is the mean primary-metric gain over gated campaign contexts; protected violation is the fraction of contexts where any protected metric exceeds tolerance; decision changed reports whether any campaign-level verdict (confirmed/null/rejected) flips relative to the default.}
\label{tab:gate}
\centering\fontsize{9}{10}\selectfont
\begin{tabular}{lccc}
\toprule
Configuration & Primary $\Delta$ $\uparrow$ & Protected viol.\ $\downarrow$ & Decision changed \\
\midrule
Full gates ($\tau = 1.5\times$, default) & +12.78 & \textbf{0.03} & -- \\
No negative-correlation gates & +14.92 & 0.41 & -- \\
All tolerances at $0.5\times$ & +11.86 & 0.04 & no \\
All tolerances at $2.0\times$ & +13.05 & \textbf{0.03} & no \\
\bottomrule
\end{tabular}
\end{table}

Table~\ref{tab:gate} shows the trade-off directly: removing the gates raises
the primary gain from $+12.78$ to $+14.92$, while protected violations rise
$13\times$. Re-calibrating all tolerances by a factor of two in either direction
changes no campaign verdict, so the result is stable across this range.

% ----------------------------------------------------------------------
\subsection{Control-Plane Ablation Under Planted Faults}
\label{app:control-plane-ablation}

H/I/E denotes hypothesis proposal, implementation, and evaluation.
Their separation keeps proposal and implementation authority distinct from
the frozen evaluation and settlement authority.

\paragraph{Control-plane audit.}
The control-plane audit complements the geometric ablations. We construct 30
trap episodes (six per type) spanning five failure modes: evaluator leakage,
within-noise gain, protected-metric sacrifice, confounded attribution, and a
single-seed fluke. Ten additional episodes contain genuine, gate-valid
improvements. Episodes are interleaved and presented without labels; each arm
settles them as confirmed, null, or rejected. We report the false-conclusion
rate and the missed-genuine rate, averaged over three seeds on Ctrl-World and
Cosmos-Predict~2.

\begin{table}[ht]
\caption{Control-plane ablation on 30 planted-fault and 10 genuine-improvement episodes. False conclusion = trap settled as \emph{confirmed}; missed genuine = valid improvement settled as \emph{null}/\emph{rejected}. Overhead is the mean extra compute per episode relative to direct settlement.}
\label{tab:control-plane}
\centering\fontsize{9}{10}\selectfont
\begin{tabular}{lccc}
\toprule
Control configuration & False conclusion $\downarrow$ & Missed genuine $\downarrow$ & Overhead (GPU-h/ep.) \\
\midrule
Full control plane & \textbf{0.07} & 0.10 & +6.2 \\
\quad w/o H/I/E separation & 0.28 & 0.09 & +4.8 \\
\quad w/o validity gates & 0.41 & 0.08 & +5.1 \\
\quad w/o progressive fidelity & 0.33 & 0.07 & $-$3.4 \\
No control plane (direct settle) & 0.52 & \textbf{0.04} & 0.0 \\
\bottomrule
\end{tabular}
\end{table}

Table~\ref{tab:control-plane} shows the trade. Direct settlement accepts nearly
every episode, but labels over half of the planted faults as confirmed effects.
The full control plane reduces false conclusions by $7.4\times$ at a cost of
6.2 GPU-h per episode, with a modest increase in missed genuine effects.
Removing H/I/E separation roughly quadruples false conclusions, especially on
confounded-attribution traps. Removing progressive fidelity saves overhead but
lets low-cost screens create confirmed records, which increases the single-seed
failure mode.

% ----------------------------------------------------------------------
\subsection{Progressive-Fidelity Screening and Selector Ablation}
\label{app:screening}

\paragraph{Progressive-fidelity screening.}
The low-cost screen should retain effective operators while saving evaluation
budget. We measure screen recall against exhaustive evaluation of all compiled
candidates, then swap the selector for simpler alternatives while holding the
screen fixed.

\begin{table}[ht]
\caption{Screening recall and selector ablation. Recall is the fraction of operators confirmed by exhaustive evaluation that the screen retains for confirmation; cost is the mean measured $C_{\mathrm{search}}$ per campaign. The exhaustive row is the recall reference, not a deployable policy.}
\label{tab:screening}
\centering\fontsize{9}{10}\selectfont
\begin{tabular}{lcc}
\toprule
Strategy & Screen recall $\uparrow$ & $C_{\mathrm{search}}$ (GPU-h) $\downarrow$ \\
\midrule
Exhaustive evaluation (reference) & 1.00 & 480 \\
Screen + UCB selector (default) & \textbf{0.94} & \textbf{205} \\
Screen + $k$NN selector & \textbf{0.94} & 214 \\
Screen + random selector & 0.91 & 268 \\
\bottomrule
\end{tabular}
\end{table}

Table~\ref{tab:screening} shows that the screen retains $94\%$ of exhaustively
confirmed operators at $43\%$ of the exhaustive evaluation cost. The $k$NN and
UCB rows have the same reported recall, with $214$ versus $205$ GPU-h. The
comparison supports screening as a source of savings and separates that saving
from the choice of acquisition rule.

% ----------------------------------------------------------------------
\subsection{Cost Ledger}
\label{app:cost-ledger}

\paragraph{Cost accounting.}
Fingerprinting and screening incur upfront cost; this ledger accounts for it per campaign. Avoided spend is estimated from the cold-start trajectory of the same campaign (main-text Figure~\ref{fig:lobo-summary}) net of measured $C_{\mathrm{overhead}}$. This ledger is separate from the main-text measured $C_{\mathrm{first+}}$.

\begin{table}[ht]
\caption{Per-campaign measured cost ledger (GPU-h). $C_{\mathrm{overhead}}$ = fingerprinting + screening + confirmation; it is reported separately from main-text $C_{\mathrm{first+}}$. Net saved = avoided spend $-$ overhead.}
\label{tab:cost-ledger}
\centering\fontsize{9}{10}\selectfont
\begin{tabular}{lccccc}
\toprule
Campaign & Fingerprinting & Screening & Confirmation & Overhead & Net saved \\
\midrule
Ctrl-World & 9 & 51 & 36 & 96 & +96 \\
Cosmos-Predict 2 & 11 & 74 & 88 & 173 & +139 \\
Cosmos3 & 8 & 60 & 62 & 130 & +132 \\
\bottomrule
\end{tabular}
\end{table}

Table~\ref{tab:cost-ledger} shows that fingerprinting costs only 8--11 GPU-h per campaign and is repaid within the first certified transfer; each listed campaign ends net-positive. Confirmation is the largest component of the listed overheads.

% ----------------------------------------------------------------------
\subsection{Budget Alignment}
\label{app:budget}

\paragraph{Budget alignment.}
Every baseline and improved variant within a campaign uses identical training steps and data, with matched nominal hyperparameter-search allowance. Measured training compute and search overhead are reported separately, so differences in outcomes cannot be attributed to different schedules or search allowances.

\begin{table}[ht]
\caption{Training-schedule alignment and measured compute for the three detailed backbone--operator comparisons.}
\label{tab:budget}
\centering\fontsize{9}{10}\selectfont
\begin{tabular}{lcccc}
\toprule
Campaign & Train steps & Data & Compute (GPU-h, base vs.\ improved) & HP trials \\
\midrule
Ctrl-World & 200k & identical & 420 vs.\ 425 & 12 \\
Cosmos-Predict 2 & 50k & identical & 380 vs.\ 386 & 10 \\
Cosmos3 & 50k & identical & 390 vs.\ 394 & 10 \\
\bottomrule
\end{tabular}
\end{table}

% ----------------------------------------------------------------------
\subsection{Selector Source Ablation}
\label{app:selection-source}
Operator-text retrieval and the frequency prior provide two additional controls
for Table~\ref{tab:selector-main}. Evaluation uses the same held-out decisions
and campaign units.

\paragraph{Complementary controls.}
These selectors test whether operator descriptions or historical success
frequency suffice to prioritize operators. Table~\ref{tab:selector-main}
compares metadata, raw responses, learned embeddings, and full fingerprints;
Table~\ref{tab:executor-control} separately removes IRG while fixing the
executor. The two ablations address the informativeness of the representation
and the contribution of IRG to the complete search procedure, respectively.

\paragraph{Uncertainty across the detailed subset.}
For the three-backbone detailed subset, full-fingerprint 95\%
intervals are [0.73, 0.90] for sign accuracy,
[3.2, 4.1] for trials, [0.02, 0.14] for negative transfer, and
[0.01, 0.09] for protected violation. These backbone-cluster bootstrap
intervals accompany the point estimates in Table~\ref{tab:selector-main}
over nine campaigns and 72 decisions, retaining the dependence among seeds
and decisions within each backbone. The full nine-target inventory is listed
in Table~\ref{tab:shared-prior-archive}; it is not replaced by this subset.

\begin{table*}[t]
  \centering
  \caption{\textbf{Additional retrieval baselines.} Only the selection signal
  changes across rows; the executor, Operator registry, nominal allowance, and
  verifier are held fixed.}
  \label{tab:selection-source}
  \fontsize{9}{10}\selectfont
  \begin{tabularx}{\textwidth}{@{}>{\raggedright\arraybackslash}X >{\centering\arraybackslash}X >{\centering\arraybackslash}X >{\centering\arraybackslash}X >{\centering\arraybackslash}X@{}}
    \toprule
    Selection source & Sign accuracy $\uparrow$ & Trials to positive $\downarrow$ &
      Neg. transfer $\downarrow$ & Protected violation $\downarrow$ \\
    \midrule
    Operator-text embedding & 0.62 & 5.8 & 0.24 & 0.16 \\
    Frequency prior & 0.57 & 5.2 & 0.29 & 0.22 \\
    \bottomrule
  \end{tabularx}
\end{table*}

% Supplementary experimental details.

\section{Additional Backbone-Specific Experiments}
\label{app:backbone-experiments}

This appendix reports experiments and diagnostics not displayed in the main
text: the eight-environment ACWM-Phys matrix, additional Ctrl-World metrics and
views, Cosmos-family operators, downstream action generation, RoboCoin
adaptation, and a probe-locality case study.

\subsection{Extended Protocol}
\label{app:extended-protocol}

We treat VERDI as a research agent for a world model and evaluate its ability
to propose operators targeted to the diagnosed failure of each
backbone. To isolate the selected operator's effect, we preserve the original
observation interface, tokenizer or VAE, decoder or renderer, data partition,
and evaluation implementation unless a selected operator explicitly targets that
component. The baseline is the original backbone before applying the operator,
compared under the same data, matched training schedule, nominal search allowance, and evaluation protocol.

The instantiated operators differ by diagnosed mechanism. Ctrl-World~\citep{guo2025ctrlworld} receives a LaMo~\citep{jiang2026lamo}-style latent-motion prior for insufficient physical consistency;
Cosmos-Predict~2~\citep{nvidia2025cosmos} receives DINOv2~\citep{oquab2024dinov2} representation guidance (REP) for structural
drift; Cosmos3 receives a combined reinforcement-learning (RL) and
self-forcing operator~\citep{huang2025selfforcing}, evaluated jointly on
long-horizon rollout and downstream control; and RoboCoin~\citep{wu2025robocoin} uses an automatically constructed data and
action-conditioning adapter. We report standard video metrics (PSNR, SSIM,
LPIPS, and FID), WorldArena-style semantic, depth, flow, smoothness, and
consistency metrics, and rollout diagnostics appropriate to each claim.
Action-conditioned adaptation additionally includes a no-action control.
Each subsection reports protected-coordinate checks alongside the gains.

\subsection{Complete ACWM-Phys Optimization Results}
\label{app:acwm-details}

Table~\ref{tab:app-acwm-repairs} provides the complete eight-environment result
behind the main-text summary. Deltas are computed as operator-applied minus baseline
under the frozen ACWM-Phys evaluator. Positive PSNR and SSIM and negative
masked MSE are favorable. The evidence column records whether a result comes
from a single checkpoint or repeated receipts.

\begin{table*}[t]
  \centering
  \caption{\textbf{Failure-specific optimization on all eight ACWM-Phys
  environments.} The selected operator differs across diagnosed failure
  families; status records the available confirmation boundary.}
  \label{tab:app-acwm-repairs}
  \fontsize{9}{10}\selectfont
  \begingroup
  \setlength{\tabcolsep}{2pt}
  \begin{tabularx}{\textwidth}{@{}>{\raggedright\arraybackslash}X >{\raggedright\arraybackslash}X >{\raggedright\arraybackslash}X >{\raggedright\arraybackslash}X@{}}
    \toprule
    Environment & Failure family & Selected operator & Checkpoint/evidence \\
    \midrule
    push\_cube & rigid contact/action & cfg schedule $1\!\to\!1.1$ & 2 receipts \\
    stack\_cube & support/contact & cfg schedule $1\!\to\!1.5$ & 2 receipts \\
    push\_rope & topology/contact & self-forcing finetune & 1 checkpoint \\
    cloth\_move & deformable memory & next forcing & 800 steps \\
    push\_sand & granular transport & mixture reweight + scale $0.5$ & 3 receipts \\
    pour\_water & fluid transport & event window + scale $0.25$ & metric pass \\
    robot\_arm & articulated dynamics & self-forcing finetune & 800 steps \\
    reacher & target-conditioned control & next forcing + scale $0.02$ & 2 receipts \\
    \bottomrule
  \end{tabularx}
  \par\medskip
  \begin{tabularx}{\textwidth}{@{}>{\raggedright\arraybackslash}X >{\centering\arraybackslash}X >{\centering\arraybackslash}X >{\centering\arraybackslash}X >{\raggedright\arraybackslash}X@{}}
    \toprule
    Environment & \makecell{$\Delta$PSNR\\$\uparrow$} & \makecell{$\Delta$SSIM\\$\uparrow$} & \makecell{$\Delta$mMSE\\$\downarrow$} & Status \\
    \midrule
    push\_cube & +0.15 & +0.0002 & -0.000284 & formal \\
    stack\_cube & +0.03 & +0.0020 & -0.000531 & formal \\
    push\_rope & +0.02 & +0.0008 & -0.000151 & formal \\
    cloth\_move & +0.94 & +0.0099 & -0.009040 & formal \\
    push\_sand & +0.64 & +0.0010 & -0.000783 & formal \\
    pour\_water & +0.85 & +0.0019 & -0.001718 & formal \\
    robot\_arm & +1.08 & +0.0098 & -0.002230 & formal \\
    reacher & +0.06 & +0.0002 & -0.000119 & formal \\
    \bottomrule
  \end{tabularx}
  \endgroup
\end{table*}

\subsection{Ctrl-World Implementation and Additional Diagnostics}
\label{app:ctrl-world}

The original Ctrl-World produces plausible short-term predictions but exhibits
motion drift, object-state jumps, and unstable long-horizon transitions. Its
diagnostic response suggests that the failure is concentrated in latent state
evolution rather than the visual encoder or renderer. VERDI therefore adds a
LaMo-style branch containing a motion prior, delta-latent predictor, and drift
readout. The objective combines reconstruction and drift consistency,
\begin{equation}
  L_{\mathrm{total}}
  = L_{\mathrm{MSE}} + \lambda_{\mathrm{drift}}L_{\mathrm{drift}},
  \label{eq:app-ctrl-objective}
\end{equation}
where $L_{\mathrm{drift}}$ aligns the predicted latent displacement with the
future motion direction.

\begin{table}[t]
  \centering
  \caption{\textbf{Additional Ctrl-World WorldArena metrics.} These three coordinates
  complement the primary metrics in Table~\ref{tab:ctrl-world-main}.
  Baseline and operator use matched data and training schedules.}
  \label{tab:app-ctrl-worldarena}
  \fontsize{9}{10}\selectfont
  \begin{tabular}{lcc}
    \toprule
    Metric & Ctrl-World & VERDI \\
    \midrule
    Aesthetic Quality $\uparrow$ & 32.75 & \textbf{36.12} \\
    Background Consistency $\uparrow$ & \textbf{86.37} & 85.11 \\
    Dynamic Degree $\uparrow$ & \textbf{45.32} & 43.78 \\
    \bottomrule
  \end{tabular}
\end{table}

\begin{table}[t]
  \centering
  \caption{\textbf{Ctrl-World video quality.} Lower is better for LPIPS and
  FID.}
  \label{tab:app-ctrl-video}
  \fontsize{9}{10}\selectfont
  \begin{tabular}{lcc}
    \toprule
    Metric & Ctrl-World & VERDI \\
    \midrule
    PSNR $\uparrow$ & \textbf{20.5099} & 20.4921 \\
    SSIM $\uparrow$ & 0.7799 & \textbf{0.7899} \\
    LPIPS $\downarrow$ & 0.1528 & \textbf{0.1432} \\
    FID $\downarrow$ & 54.0076 & \textbf{51.7057} \\
    \bottomrule
  \end{tabular}
\end{table}

These additional diagnostics show the trade-off of the motion operator:
background consistency drops by 1.26 and dynamic degree by 1.54, while PSNR is
approximately flat. The background regression remains within tolerance;
dynamic degree exceeds tolerance on the open-drawer context group. That
violation prevents a verified positive for a contract covering that group.
The coordinate improvements therefore remain a descriptive trade-off for the
current contract. A narrower claim would require a scope declared before
evaluation or a fresh confirmation under a new frozen contract.
The multi-view example below tests a different task from
Figure~\ref{fig:ctrl-world-main}.

\begin{figure}[H]
    \centering
    \includegraphics[width=0.85\textwidth]{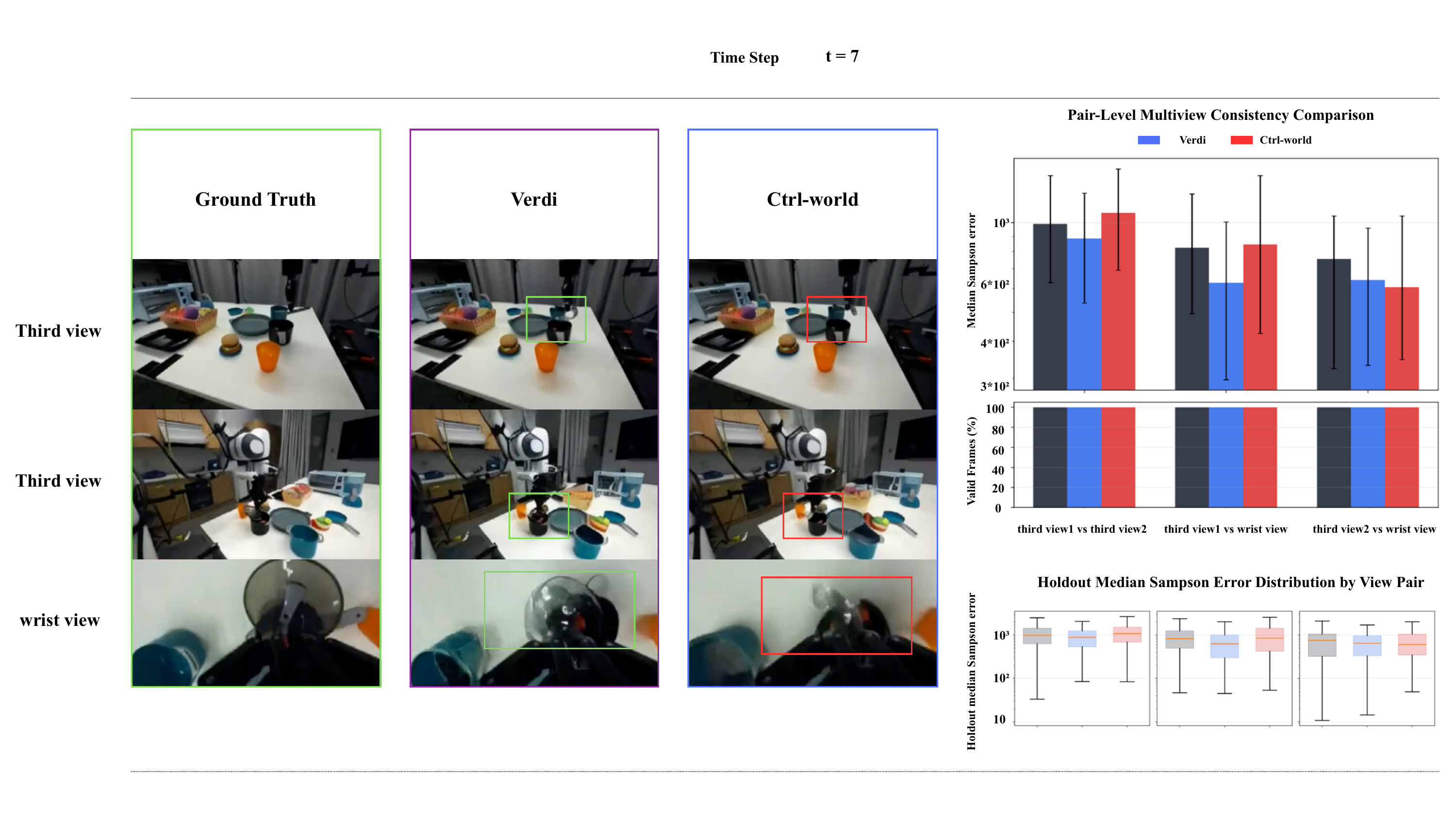}
    \caption{Consistency comparison across three camera views on the ``put
    the lid on'' task (DROID~\citep{khazatsky2024droid}). Two views improve significantly over the
    Ctrl-World baseline; third-person view~2 and the wrist view degrade,
    consistent with the validity-gate analysis above.}
    \label{fig:image2}
\end{figure}

\subsection{Structure and Long-Horizon Generation in the Cosmos Family}
\label{app:cosmos-family}

The Cosmos family exposes two different failure modes. Cosmos-Predict~2 shows
structural drift under complex motion and occlusion, whereas Cosmos3 shows
train--inference mismatch and compounding error during long rollouts. VERDI
therefore selects representation guidance for the former and a combined RL +
self-forcing operator for the latter rather than applying one global recipe.

\paragraph{Cosmos-Predict~2: representation-guided prediction.}
REP injects DINO features as an additional constraint so that prediction is
aligned not only in pixel space but also in a higher-level structural feature
space. The complete eight-coordinate WorldArena result is retained here in
Table~\ref{tab:app-cosmos-rep}, while the main text uses Cosmos3 for the
complementary long-horizon and control study.

\begin{table}[t]
  \centering
  \caption{\textbf{Cosmos-Predict~2 results on WorldArena.} Baseline versus
  REP-guided operator under matched training schedules; measured compute is reported separately.}
  \label{tab:app-cosmos-rep}
  \fontsize{9}{10}\selectfont
  \begin{tabular}{lcc}
    \toprule
    Metric & Cosmos-Predict~2 & VERDI \\
    \midrule
    Semantic Alignment $\uparrow$ & 90.38 & \textbf{90.82} \\
    Depth Accuracy $\uparrow$ & 95.35 & \textbf{97.05} \\
    Aesthetic Quality $\uparrow$ & 36.05 & \textbf{37.30} \\
    Background Consistency $\uparrow$ & \textbf{85.35} & 82.73 \\
    Dynamic Degree $\uparrow$ & 42.18 & \textbf{43.15} \\
    Flow Score $\uparrow$ & 29.55 & \textbf{30.70} \\
    Motion Smoothness $\uparrow$ & 78.93 & \textbf{81.31} \\
    Subject Consistency $\uparrow$ & \textbf{83.07} & 81.79 \\
    \bottomrule
  \end{tabular}
\end{table}

Background consistency drops by 2.62 and subject consistency by 1.28. The
background regression exceeds its tolerance on the BridgeData V2~\citep{walke2023bridge}
context group, so the current contract records the REP result as a mixed
outcome. The archive keeps the regression together with the improved
coordinates; a narrower claim would require a separately declared contract.

\begin{figure}[H]
    \centering
    \includegraphics[width=0.85\textwidth]{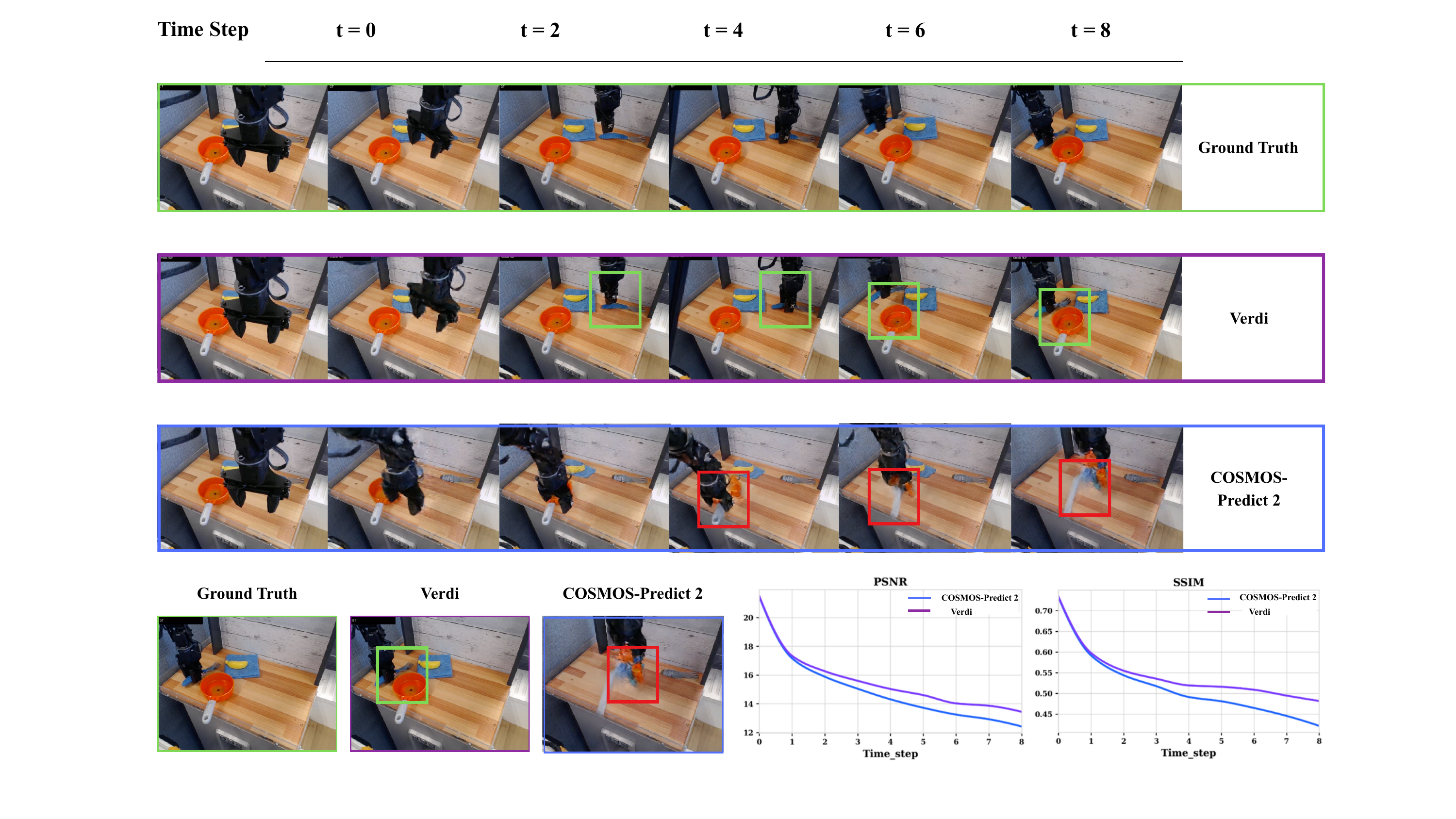}
    \caption{Qualitative comparison of Cosmos-Predict 2 and REP on ``place
    the fork in the pot'' (BridgeData): ground truth, baseline, and
    REP-improved prediction.}
    \label{fig:image3}
\end{figure}

\paragraph{Cosmos3: RL + self-forcing.}
Cosmos3 performs well in short rollouts but shows subject and background
drift as generated states become its own context. The RL component targets
downstream action outcomes, while self-forcing exposes the model to generated
histories during training while preserving its architecture and input--output
interface. Together they target action quality and the diagnosed
train--inference mismatch.

Figure~\ref{fig:cosmos3-long-horizon}b shows the resulting long-horizon
generation on ``loosen the pot handle'' (DROID~\citep{khazatsky2024droid}).
Action-level diagnostics are reported in App.~\ref{app:robot-evaluation}.

\subsection{Simulation and Additional Action Diagnostics}
\label{app:robot-evaluation}

The physical-robot evaluation is reported in \S\ref{sec:exp-real-robot} and
Figure~\ref{fig:real-robot}. Here we examine the complementary Franka simulation
setting: pick-and-place, drawer opening, and peg insertion. Each seed uses 50
episodes per arm and task; the table reports the mean success percentage across
the evaluated seeds, so entries need not be multiples of two percentage points.
Vanilla and Cosmos3 with the selected operator use eight-step action chunks with
re-observation after each chunk.

\begin{table}[t]
  \centering
  \caption{\textbf{Simulated closed-loop manipulation success rate.} Vanilla
  versus Cosmos3 with the combined RL + self-forcing operator on paired
  simulation initial states. Each seed has 50 episodes per arm and task; entries
  are means across seeds and $\Delta$ is in percentage points.}
  \label{tab:app-robot-success}
  \fontsize{9}{10}\selectfont
  \begingroup
  \begin{tabular}{llcccc}
    \toprule
    Environment & Task & Avg. steps & Vanilla Cosmos3 & With operator & $\Delta$ \\
    \midrule
    Sim (50/seed) & Pick-and-place & 8  & $62\%$ & $\mathbf{74\%}$ & $+12.0$ \\
    Sim (50/seed) & Drawer opening & 11 & $41\%$ & $\mathbf{59\%}$ & $+18.0$ \\
    Sim (50/seed) & Peg insertion  & 14 & $28\%$ & $\mathbf{52\%}$ & $+24.0$ \\
    \bottomrule
  \end{tabular}%
  \endgroup
\end{table}

\begin{table}[t]
  \centering
  \caption{\textbf{Action-level metrics.} Vanilla versus RL + self-forcing
  Cosmos3 on expert-action L2 error and the descriptive success gap between
  the 8-step and 14-step tasks. The gap mixes task and step-count differences.}
  \label{tab:app-robot-action}
  \fontsize{9}{10}\selectfont
  \begin{tabular}{lcc}
    \toprule
    Metric & Vanilla Cosmos3 & With operator \\
    \midrule
    Expert-action L2 error $\downarrow$ & $0.084$ & $\mathbf{0.061}$ \\
    8-to-14-step task gap $\downarrow$ & $-34$ pts & $\mathbf{-22}$ pts \\
    \bottomrule
  \end{tabular}
  \end{table}

Applying the selected operator improves average simulation success by 18.0 percentage
points in this three-task comparison. Expert-action L2 error decreases from
0.084 to 0.061, while the descriptive 8-to-14-step task gap narrows from 34 to
22 points. Because the two rows use different tasks as well as different step
counts, this gap is not interpreted as a standalone estimate of error
accumulation with horizon.

\subsection{Bimanual AgileX Cobot Magic Qualitative Evidence}
\label{app:songling}

We additionally show representative paired rollouts on an AgileX Cobot Magic dual-arm
robot. We conducted 40 validations for 3 task types, including "use of left and right hands", "pick and place", and "cooperation of left and right hands". As a result, the task success rate of VERDI increased by 19.6 percentage points. Red boxes mark
the baseline's failed or misaligned interaction states, while green boxes mark
the corresponding successful interaction states produced with the VERDI operator.

\begin{figure}[H]
  \centering
  \includegraphics[width=0.98\textwidth]{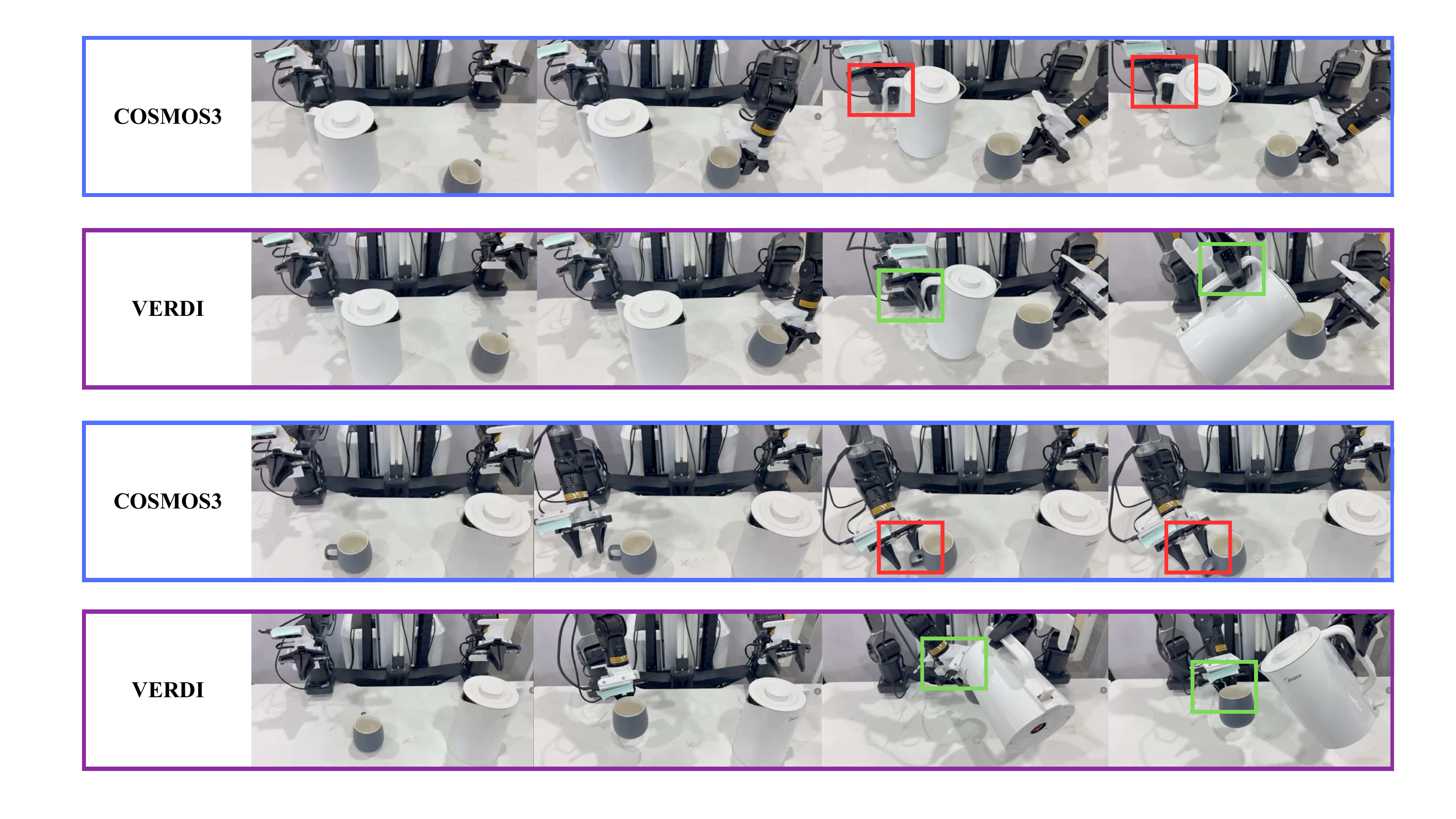}
  \caption{\textbf{Qualitative AgileX Cobot Magic dual-arm evaluation.} Representative
  paired frames for two manipulation sequences. The top and third rows show
  Cosmos3 baseline rollouts; the second and fourth rows show VERDI rollouts.
  Colored boxes identify the interaction state at the corresponding stage of
  each sequence.}
  \label{fig:songling}
\end{figure}

\subsection{Automatic Adaptation to RoboCoin}
\label{app:robocoin}

RoboCoin tests a different capability: converting a dataset with a new robot
embodiment, action interface, camera layout, and task distribution into an
executable world-model training task. VERDI parses episode boundaries,
observations, action sequences, camera identities, temporal alignment, and
embodiment metadata, then constructs a unified training and replay contract.
It selects the DreamDojo/Cosmos-Predict~2.5 action-conditioned model initialized
from AgiBot post-training weights and materializes the field mapping,
action-conditioning, temporal sampling, and camera organization.

For single-view adaptation, the video stream is aligned with robot actions to
form an action-conditioned future-video prediction task. The no-action control
tests whether the prediction actually depends on the action input. Action
conditioning improves held-out LPIPS by 0.029 and SSIM by
0.039 relative to this control. For multi-view adaptation, synchronized camera
identities are retained rather than treating views as interchangeable; the
reported cross-view subject-consistency variance is 0.029 versus 0.083 for
independent per-view adaptation.

\begin{figure}[H]
    \centering
    \includegraphics[width=0.75\textwidth]{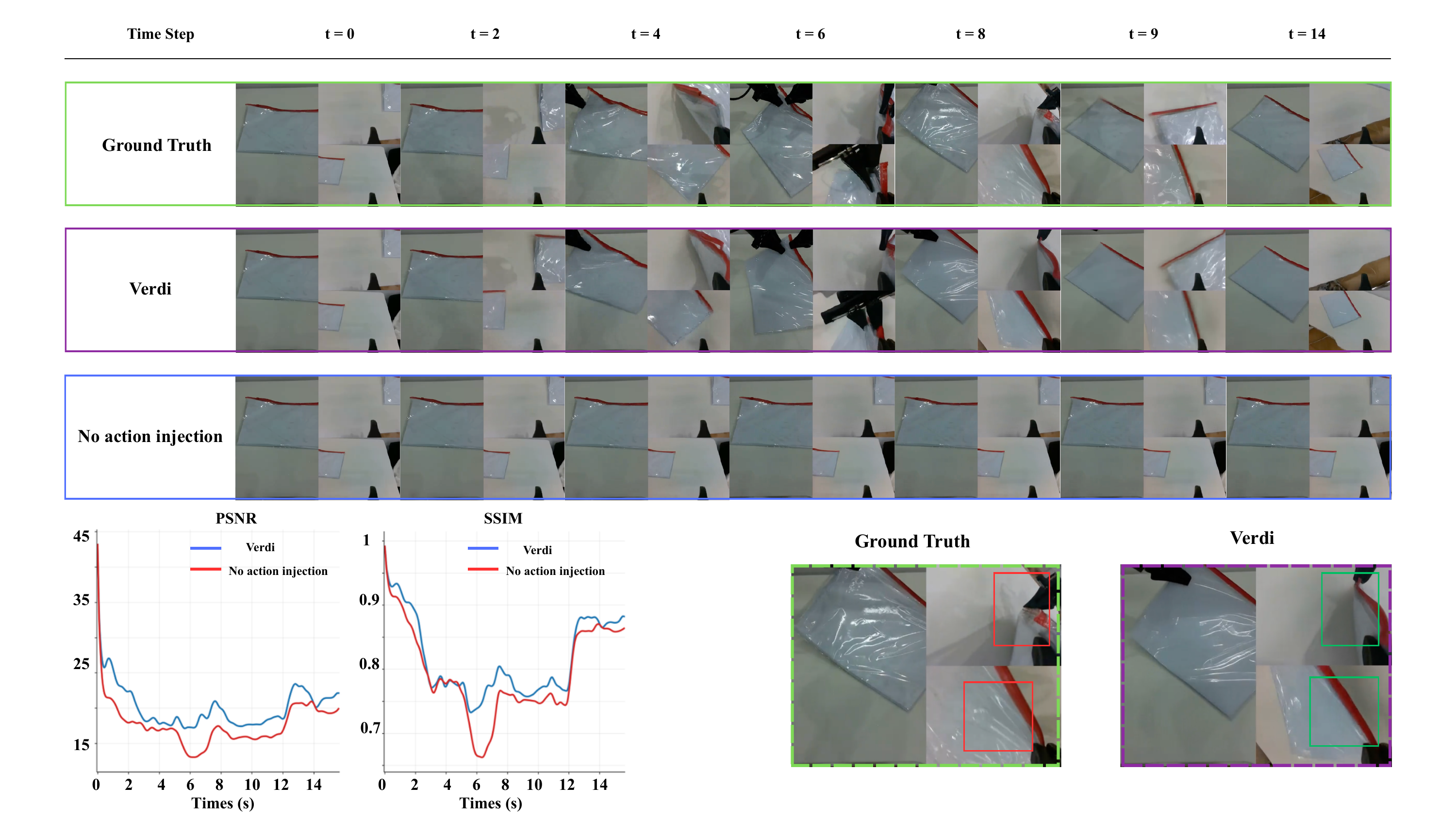}
    \caption{Action-conditioned adaptation on RoboCoin (``opening the red
    folder''): ground-truth future, action-conditioned prediction, and
    no-action control, organized in the same camera order and time frame
    across all three synchronized views. The gap between the
    action-conditioned and no-action rows evidences that the model uses the
    action input rather than unconditional visual continuation.}
    \label{fig:image5}
\end{figure}

\begin{figure}[H]
\centering
\includegraphics[width=0.86\textwidth]{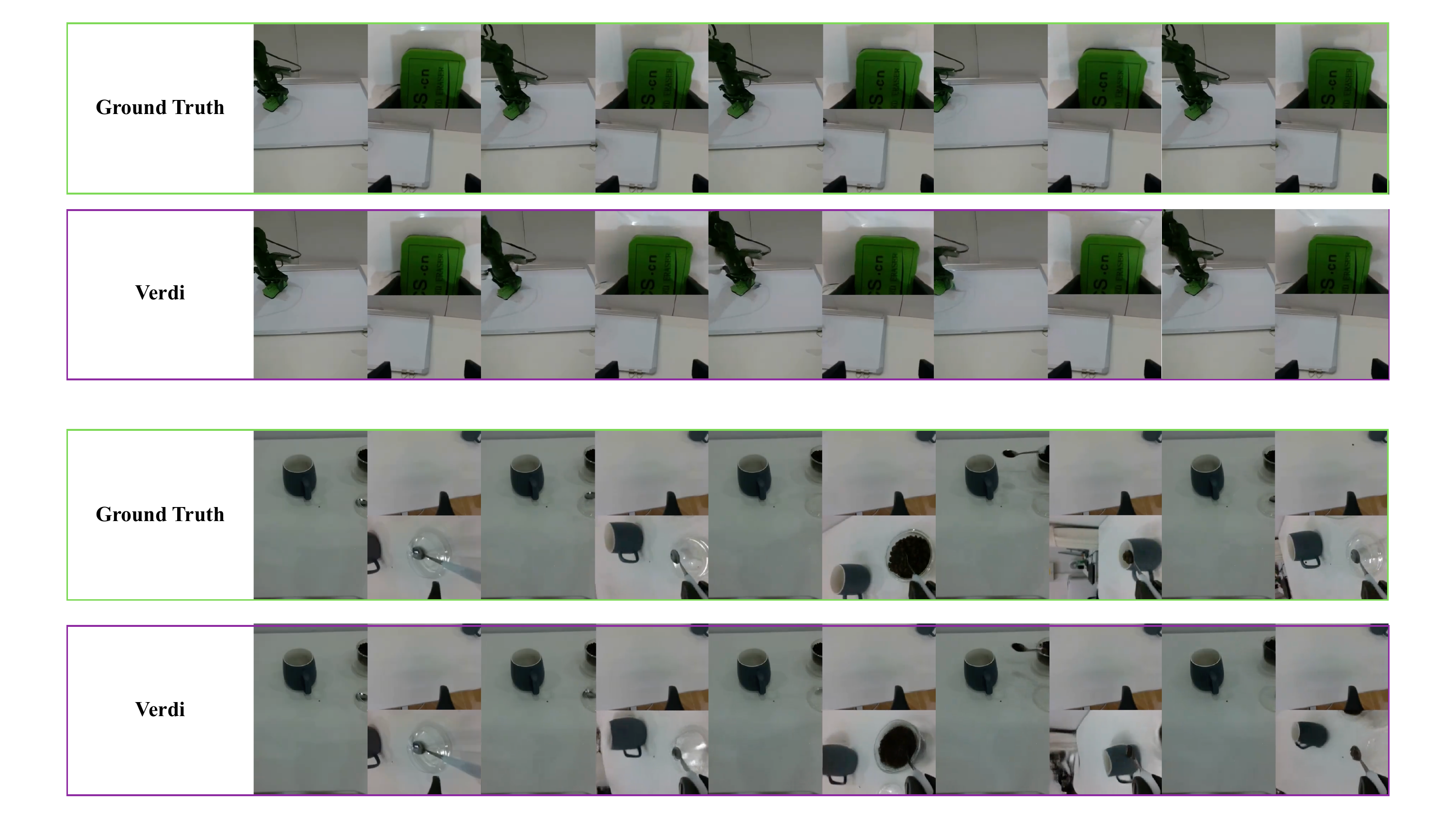}
\caption{\textbf{Ground-truth vs.\ predicted frames on two RoboCoin tasks}
(box relocation, top pair; cup pouring, bottom pair; two synchronized views
per time step). VERDI-adapted Cosmos-Predict~2.5 (purple) tracks the ground
truth (green) closely on both tasks, including object identity and
multi-view correspondence.}
\label{fig:robocoin-qual}
\end{figure}

This experiment gives one working adaptation contract for a single dataset
family and base model. Broader embodiments and datasets remain future tests.

\vspace{0.5em}
\subsection{Probe Locality: A Counterexample Case Study}
\label{app:qualitative-main}

\begin{figure}[H]
  \centering
  \includegraphics[width=0.85\columnwidth]{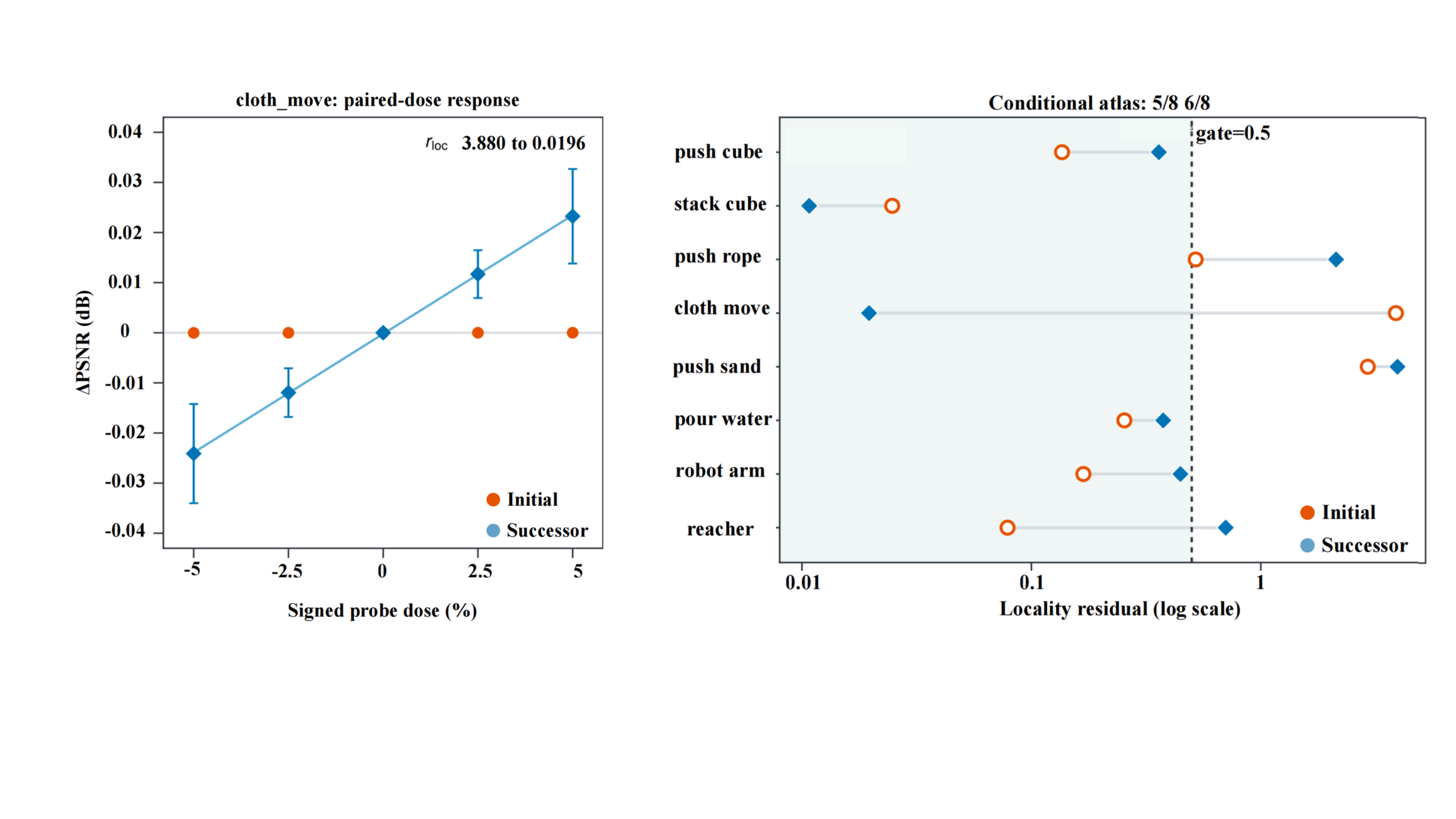}
  \caption{\textbf{Counterexample-driven probe evolution.} A non-local
  response triggers a temporal-mix probe. The library keeps both
  charts and admits the successor only where its paired-dose locality gate
  passes.}
  \label{fig:probe-evolution}
\end{figure}

This case tests local chart validity, separately from the nested operator-selection regret
replay in \S\ref{sec:exp-probe-evolution}. The initial probe is local in five of
eight environments. A cloth-move counterexample triggers an action-embedding
temporal-mix probe, reducing the locality residual from $3.88$ to $0.0196$ and
increasing held-out chart coverage from 5/8 to 6/8
(Figure~\ref{fig:probe-evolution}). The result is a conditional chart-coverage
improvement; downstream operator gains are evaluated separately in the frozen
selector replay.

\section{Limitations}
\label{sec:limitations}

Although we validate \methodname{} on nine backbones with multiple random seeds, and further confirm its effectiveness in both real-world and simulated environments, our evaluation is constrained by limited human and computational resources. The nine backbones were chosen as representative, mainstream architectures rather than through an exhaustive enumeration of all possible backbones, and thus do not cover the majority of existing world models. In future work, we plan to collaborate with the community to extend our experiments and analysis to a broader range of backbone architectures.

\end{document}